\documentclass[times,twocolumn,review,nopreprintline]{elsarticle}
\usepackage{caption} 
\usepackage{medima}
\usepackage{framed,multirow}

\usepackage{amssymb}
\usepackage{amsmath}
\usepackage{latexsym}

\usepackage{color, soul}
\usepackage{xargs}
\usepackage{tikz}
\usepackage{todonotes}
\usepackage{multirow} 
\usepackage{verbatim}
\usepackage{nameref,hyperref}
\usepackage{url}
\usepackage{xcolor}

\usepackage{hyperref}

\definecolor{newcolor}{rgb}{.8,.349,.1}

\journal{Medical Image Analysis}

\usepackage{comment}

\usepackage{fancyhdr}		%Gives headers and footers defined below
\usepackage[section]{placeins}   %
\usepackage{subcaption}
\usepackage{url}
\usepackage{caption} 
\usepackage{framed,multirow}

\usepackage{amssymb}
\usepackage{amsmath}
\usepackage{latexsym}

\usepackage{color, soul}
\usepackage{xargs}
\usepackage{tikz}
\usepackage{todonotes}
\usepackage{multirow} 
\usepackage{verbatim}
\usepackage{nameref,hyperref}

\usepackage{url}
\usepackage{xcolor}

\usepackage{hyperref}

\usepackage{graphicx}

\makeatletter \renewcommand\@biblabel[1]{$^{#1}$} \makeatother
\newcommand{\cen}[1]{\begin{center} #1 \end{center}}

\renewcommand{\refname}{}       %

\usepackage{hyperref}
\hypersetup{ colorlinks,
    citecolor=blue,
    filecolor=blue,
    linkcolor=blue,
    urlcolor=blue
}

\usepackage{xcolor}
\definecolor{gray}{rgb}{0.6,0.6,0.6}
\definecolor{red}{rgb}{0.85,0,0}
\definecolor{green}{rgb}{0,0.85,0}
\definecolor{blue}{rgb}{0,0,0.85}
\definecolor{beige}{rgb}{0.92,0.87,0.78}
\usepackage[all]{hypcap}    %causes link to figures to go to figure, not caption
\begin{document}
%\verso{Harini Veeraraghavan \textit{et~al.}}
\begin{frontmatter}
\title{Nested-block self-attention for robust radiotherapy planning segmentation}
		%Cross-modality (MR-CT) educed deep learning (CMEDL) for segmentation of lung tumors on CT
		%or include affiliations in footnotes: % \ead[url]{www.elsevier.com}
		\author[mymainaddress]{Harini Veeraraghavan\corref{mycorrespondingauthor}$^{\dagger}$}
		\cortext[mycorrespondingauthor]{Corresponding author: Harini Veeraraghavan, Memorial Sloan-Kettering Cancer Center, Medical Physics Department,  1275 York Ave. New York NY 10065,
			Telephone: (646)888-8015.\\
			$^{\dagger}$ Equal contribution}
		\ead{veerarah@mskcc.org }
		\author[mymainaddress]{Jue Jiang$^{\dagger}$}
		\author[mymainaddress]{Sharif Elguindi}
		\author[mymainaddress]{Sean L. Berry} 
		\author[mysecaddress]{Ifeanyirochukwu Onochie} 
		\author[mymainaddress]{Aditya Apte} 
		\author[mymainaddress]{Laura Cervino} 
		\author[mymainaddress]{Joseph O. Deasy }
		
		\address[mymainaddress]{Department of Medical Physics,Memorial Sloan-Kettering Cancer Center, New York, USA.} 
		\address[mysecaddress]{Department of Radiation Oncology, Memorial Sloan-Kettering Cancer Center, New York, USA.}
		
\begin{abstract}
%Delivering conformally high doses to the target, while reducing risk of unnecessary radiation to the nearby normal organs in the head and neck (H&N) requires accurate segmentation of a large number of organs at risk (OAR). 
Although deep convolutional networks have been widely studied for head and neck (HN) organs at risk (OAR) segmentation, their use for routine clinical treatment planning is limited by a lack of robustness to imaging artifacts, low soft tissue contrast on CT, and the presence of abnormal anatomy.  In order to address these challenges, we developed a computationally efficient nested block self-attention (NBSA) method that can be combined with any convolutional network. Our method achieves computational efficiency by performing non-local calculations within memory blocks of fixed spatial extent. Contextual dependencies are captured by passing information in a raster scan order between blocks, as well as through a second attention layer that causes bi-directional attention flow. We implemented our approach on three different networks to demonstrate feasibility. Following training using 200 cases, we performed comprehensive evaluations using conventional and clinical metrics on a separate set of 172 test scans sourced from external and internal institution datasets without any exclusion criteria. NBSA required a similar number of computations (15.7 gflops) as the most efficient criss-cross attention (CCA) method and generated significantly more accurate segmentations for brain stem (Dice of 0.89 vs. 0.86) and parotid glands (0.86 vs. 0.84) than CCA. NBSA's segmentations were less variable than multiple 3D methods, including for small organs with low soft-tissue contrast such as the submandibular glands (surface Dice of 0.90). Our method has been in routine clinical use since May, 2020. It reduced contouring effort by 61\% compared to manual contouring and the dose distribution metrics derived from our method were strongly correlated ($R^2 > 0.9$, $r=1$) with those derived from  manual segmentations. In conclusion, we developed and clinically implemented a new self-attention based OAR segmentation method that is robust to routinely seen clinical conditions for HN radiation therapy treatments.
\end{abstract}


	\end{frontmatter}

\section{Introduction}
\begin{figure*}
\centering
\includegraphics[width=0.7\textwidth,scale=0.3]{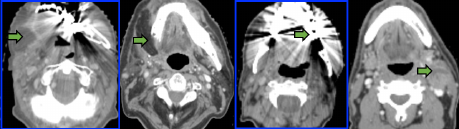}
\caption{\small {Examples of challenging cases: tumors that deform anatomy combined with dental artifacts, surgically excised mandible and tumor, severe artifacts, and an enlarged lymph node abutting the contralateral submandigular gland. }}
\label{fig:DifficultCases}
\end{figure*}
Highly conformal image guided radiation treatments such as intensity modulated radiation therapy (IMRT) have shown to improve both treatment outcomes and patients' quality of life in head and neck (HN) cancer patients by sparing normal organs of unnecessary radiation \citep{Harari2010}. But these treatments require accurate organ at risk (OAR) segmentation. Complexity of the OAR morphology due to the large number of organs occupying HN, a lack of sufficient soft-tissue contrast that clearly identifies organ boundaries, and difficult but common imaging conditions such as dental artifacts and invading tumors all make HN OAR segmentation a difficult task\citep{ibragimov2017}. Manual delineation of  HN OARs is highly time consuming\citep{Harari2010,vandijk2020} taking upto 2 hours. Manual segmentations also exhibit high inter-observer variability\citep{nelms2010}. This has motivated the development of both atlas-based\citep{fritscher2016deep,duc2015,haq2019} and deep learning methods\citep{ibragimov2017,nikolov2018deep,zhu2018anatomynet,liang2020,wang2018hierarchical}. 
\\
An important requirement for routine clinical use of deep learning methods in radiation therapy planning is the robustness of these methods to the commonly seen difficult clinical conditions, which include moderate to severe dental artifacts and abnormal anatomy. Severe dental artifacts can completely obscure glandular organs\citep{ibragimov2017}. Surgically excised organs and enlarged targets can deform the adjacent normal organs, which can adversely impact automated methods.  
\\
Prior works on HN organ segmentation have predominantly focused on improving segmentation accuracy under low soft-tissue contrast and varying organ sizes. The most common strategy is a multi-step method to first localize the organs followed by organ-specific pixel-wise segmentation\citep{ibragimov2017,wang2018hierarchical,men2019,ren18,liang2020,chen2020}. These methods try to focus the network towards relevant regions containing the organs either by explicitly localizing the organs\citep{wang2018hierarchical,men2019,liang2020} or by using a shape constraint pertaining to that organ\citep{tong2018}. Squeeze and excitation methods\citep{zhu2018anatomynet} identify relevant feature maps while image-level attention gating mechanisms use organ segmentation probabilistic maps to refine segmentations\citep{WangMedIA2019,oktay2018}. 
\\
However, the basic assumption of these methods is that the organs are visible in their usual location despite low soft-tissue contrast. This assumption is violated when large tumors abutting the organs deform and push them out of their usual location, organ appearances are severely altered due to surgical excision of diseased organs, and severe artifacts entirely obscure and alter the local statistics of an organ (see Figure~\ref{fig:DifficultCases}). In these situations, the afore-mentioned methods can no longer operate effectively as the first-stage of organ localization may itself fail. The unusual organ statistics may also preclude robust algorithmic inference. Therefore, we developed a new self-attention based approach to handle such difficult yet common imaging conditions.
\\

\begin{figure*}
\centering
\includegraphics[width=0.8\textwidth,scale=0.2]{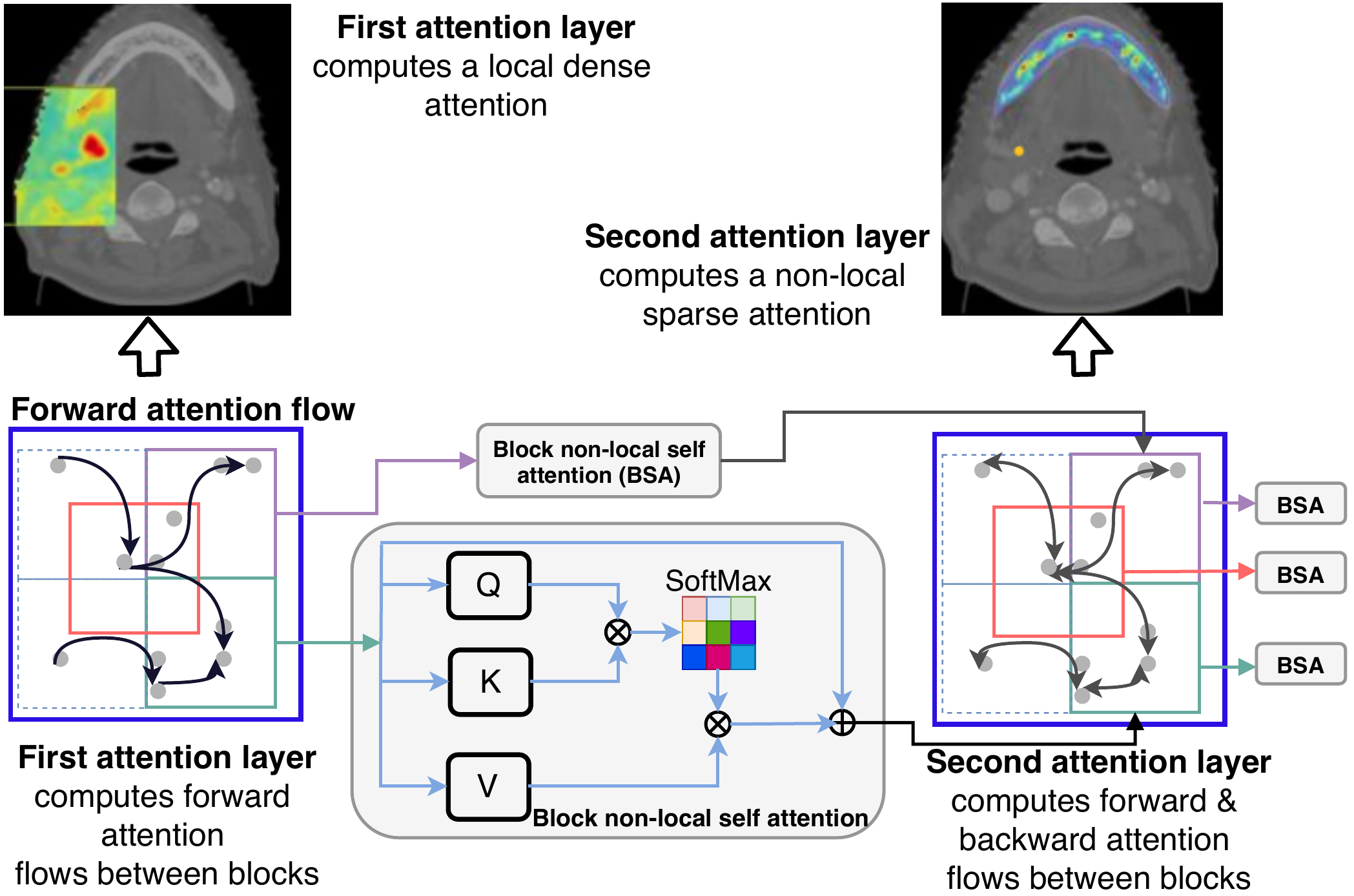}
\caption{\small {Schematic description of the nested block self-attention method. The first attention layer computes non-local attention within the individual memory blocks. Information is passed on to the neighboring memory blocks in a raster scan order. The second attention layer computes the non-local attention using the results from the first layer, which causes a backward flow of information, leading to bi-directional self-attention computation.}}
\label{fig:block}
\end{figure*}

Our approach, called nested block self-attention (NBSA), combines interactions between regional contexts, while extracting the non-local attention information in a fast and memory efficient way. The key idea is to split an image into fixed local 2D memory blocks of defined spatial extent and compute a non-local self-attention in each block in a raster scan order. The memory blocks are placed overlapping each other's context to allow the flow of attention from one block to another. The outputs for all the blocks are then combined with the original image to produce a local self-attention map. A second NBSA attention layer is added to allow bi-directional attention flow (Figure.~\ref{fig:block}). By restricting the non-local computations to the memory blocks, the memory and computational needs of our method are similar to the criss-cross self attention or CCA\citep{huang2019ccnet}, the most memory efficient method published to date. Our method also captures more long-range contextual information than the CCA by including off-diagonal pixels and passing information between the regional contexts or memory blocks. 
\\
Notably, we have been using this method for routine clinical treatment planning at our institution since May, 2020 and have already analyzed close to 1000 cases. To our knowledge, this is the first report of a deep learning method used for routine treatments. Whereas commercial developments in deep learning methods are underway, we are not aware of reports of detailed clinical evaluation of such methods on routine clinical data. 
\\
Our contributions are: 
\begin{itemize}
    \item \textit{New self-attention method: \/}\rm We introduced a computationally and memory efficient nested block self attention method for modeling both local and long-range context interactions.
    \item \textit{First deep learning application for routine clinical treatment planning: \/}\rm To our best knowledge, this is the first report of a deep learning model applied for routine clinical treatment planning for head and neck cancers. Importantly, we present results of analyzing clinical image sets with complex anatomical conditions and without any exclusion criteria. 
    \item \textit {Implementation on multiple network architectures: \/}\rm We present results of implementing the NBSA on three different segmentation architectures with varying number of parameters. 
    \item \textit{Rigorous evaluation: \/}\rm We present results of evaluating our method on routine and prospectively acquired clinical cases using geometric and clinical utility metrics. We also evaluated our method against related self-attention methods using geometric accuracy and computational requirements. We compared the geometric accuracy of our method against 3D networks, computed robustness to inter-rater segmentations, and conducted experiments to study impact of network design. 
\end{itemize} 

\section{Related Works}
\subsection{HN OAR segmentation methods}
 Numerous works have employed single and multi-atlas based methods for generating OAR segmentations\citep{fritscher2016deep,duc2015,haq2019}. Deep learning methods\citep{ibragimov2017,wang2018hierarchical,zhu2018anatomynet,nikolov2018deep} have shown as good or better performance than atlas based methods with minimal testing time computational requirements. An extensive recent review of head and neck segmentation methods is in\citep{vrtovec2020}.
 \\
Previous approaches either use separate sub-networks trained to segment the individual organs\citep{ibragimov2017,wang2018hierarchical,men2019,ren18,vandijk2020} or a single network to simultaneously segment all the organs \citep{liang2020,chen2020,tong2018}. Hierarchical methods have used step-wise shape constraints, wherein the large and easy to locate organs are used to constrain small organs with low soft-tissue contrast\citep{wang2018hierarchical}. A different approach consists of extracting a latent embedding of organ shapes to regularize segmentation network training\citep{tong2018}. Hierarchical constraints, whereby segmentation of organs with higher contrast (e.g. mandible) are used as additional inputs to constrain the segmentation of moderately hard (low-soft tissue contrast but large organs like the parotids) followed by small and hard organs (e.g. optic nerves and brachial plexus) was recently introduced with neural architecture search for segmenting up to forty-two different HN organs\citep{guo2020}. Liu et.al\citep{liu2020} employed a different strategy, wherein a generational adversarial network was used to transform CT into pseudo MRI images, on which segmentations were generated. 
\\
An alternative strategy is to focus the network's attention towards relevant regions in the image. Squeeze and excitation blocks are used in the  Anatomynet\citep{zhu2018anatomynet} to compress global feature channels and extract relevant features. Additionally, this method reduced the number of down-sampling calculations to have sufficient information to segment small organs in the reduced size feature maps. Separate branches with shared feature layers were used by focusNet\citep{gao2019MICCAI} to segment multiple OARs. Organ-specific gating techniques combine the advantages of a multi-step approach with focused attention approach, wherein a probabilistic segmentation indicative of the likelihood of the organs locations in the image from the first stage are used to further refine the segmentation in the second stage\citep{WangMedIA2019,oktay2018}. 
 
\subsection{Self-attention methods}
Self-attention (SA)\citep{vaswani2017attention,parmar2018image} is more effective than CNNs for handling complex imaging conditions because it can model both local and long-range anatomical contexts. CNNs rely on fixed kernel size structure to capture local dependency and the recursive pooling to capture longer range context. But this is inherently limiting as several layers are required and important image resolution information is lost due to pooling operations done to capture the long range dependencies\citep{Chen2018A2NetsDA}. Standard non-local SA\citep{wang2018non}, which is related to computing non-local means has shown to be highly effective in improving the accuracy of image-classification and object detection tasks. Point-wise self-attention\citep{zhao2018psanet} is a type of SA that models the long-range interactions by aggregating features using feature weights distributed to different locations after computing point-wise interactions between all feature-pairs in the image. 
\\
However, SA requires memory intensive calculations, which limits the size of images that can be used\citep{ramachandran2019standaloneSA}. Prior works attempted to reduce the computational burden by modeling only the correlations between distinct objects detected in an image~\citep{hu2018relation}, by successive pooling~\citep{yuan2018ocnet}, as well as by restricting the self-attention calculation to a fixed spatial extent\citep{ramachandran2019standaloneSA,shen2018bi}. A different form of SA was proposed as a product of two sparse affinity matrices that quantify the level of interactions between the long and short range contexts~\citep{Huang2019InterlacedSS}. But this approach relies on permutation of voxels to compute the long and short-range dependencies, thereby creating a tradeoff between the number of permutations done to capture the dependencies vs. the computations. Also, the choice of permutation is highly application specific. 
\\
The idea that SA is fundamentally a way of biasing the network computations towards more informative parts of an image or text sequence has been extended to extract relevant features through lightweight gating mechanisms like the squeeze and excitation (SE)\citep{Hu2020SqueezeExcitation} and attention-gated methods\citep{oktay2018}. A hybrid approach combining both spatial and feature channel aggregation was used by Fu et.al\citep{fu2018dual} for semantic segmentaiton of natural images. SE methods are particularly interesting in that SE blocks are segmentation architecture independent computational units that enhance the representational power of the network by modeling channel-wise dependencies. SE blocks have been used for handling different organ sizes in HN\citep{zhu2018anatomynet,gao2019MICCAI} and other organ segmentations\citep{roy2018concurrent}. However, the global average pooling used in the SE blocks weights the contributions of all features equally and ignores the differential relevance of the different sets of features to inform a particular context. The issue of global average pooling was mitigated in the double attention networks\citep{Chen2018A2NetsDA} that used a transposed multiplication of the attention maps to weight the interactions. This transposed multiplication is in a sense like computing the eigen values of the attentions such that only the most relevant long-range and short-range interactions for each feature are used in an adaptive manner. 
\\
An even more efficient and domain independent approach to mitigate the memory and computational needs for self-attention is the criss-cross self-attention (or CCA)\citep{huang2019ccnet} that considers only the horizontal and vertical elements as the contextual neighborhood to compute the second order self-attention pooling. This method has also shown to be successful for lung segmentation from chest X-ray images\citep{Tang2019CCNet}. However, CCA makes an overly simplistic assumption that all the relevant long-range information resides in the horizontal and vertical neighborhoods. 
\\
Our method is \textcolor{black}{related} to the masked or bi-directional block self-attention method\citep{shen2018bi} used for sentence modeling and the stand-alone self-attention\citep{ramachandran2019standaloneSA} used for object classification, in that all three methods use memory blocks of fixed spatial extent. But there are fundamental differences. The standalone attention uses a pixel-centered blocks with relative position weighting. The attention layers are also used within the convolutional stem before the pooling operations as a way of obtaining weighted average pooling. The bidirectional block self-attention\citep{shen2018bi} also computes a bi-directional attention flow between the memory blocks like ours. However, the attention computation requires a second order pooling of the attention features from the first intra-block attention layer, which are then unpooled as weighted feature weights following the second masked attention and then fused with the top-level features. Our method on the other hand is not an approach to compute weighted pooling. It is instead a way of computing the correlation structure of the various elements in an image and extract the local and long-range dependencies. Also, the attention layers use an identical structure for simpler implementation as a network independent attention block. To our best knowledge, ours is the first to use the nested attention method for segmenting multiple organs with varying sizes, low soft-tissue contrast, and other difficult imaging conditions.

\section{Method}
\label{sec:method}
The nested block self-attention (NBSA) method consists of two consecutively placed attention layers to extract the bi-directional attention flow between all parts of the image. A NBSA attention layer is implemented by placing multiple spatially distinct memory blocks covering the entire image. The memory blocks are similar to the multi-head attention blocks\citep{shen2018bi}. In order to achieve memory efficiency, the transformation weights for self-attention computation are shared across all the blocks. Multi-head attention can be obtained by using different transformation weights.

\subsection{Non-local self attention}
The self attention scheme performs a non-local operation to compute all-pairs affinity for each pixel in a given image. Let a $X \in {\Bbb R}^{N \times C}$ be a feature map computed through a convolutional operation on an input image $I$, where $N = H \times W$ for a 2D and $N = H \times W \times Z$ for a 3D image, with H, W and Z being the height, width and number of slices in the image, and C is the number of feature channels. The non-local attention is computed as a weighted sum of features at all positions, with the weights corresponding to the correlation of the features, 
\begin{equation}
\label{eqn:SA}
\begin{split}
\boldsymbol{\beta} = f(\theta(\boldsymbol{X}), \phi(\boldsymbol{X}))g(\boldsymbol{X}), \textrm{where}, \quad 
\theta(\boldsymbol{X}) = X\boldsymbol{W_{\theta}}, \, & \\ \phi(\boldsymbol{X}) = X\boldsymbol{W_{\phi}}, \, g(\boldsymbol{X}) = X\boldsymbol{W_{g}}, 
\end{split}
\end{equation}
where $\theta(.), \phi(.), g(.)$ are learnable transformations of the input feature map \textbf{X} corresponding to the query $\textbf{Q} = \theta(X)$, key $\textbf{K} = \phi(X)$, and value $\textbf{V} = g(X)$. These feature transformations are often computed using a $1 \times 1$ convolution with weight matrices $\boldsymbol{W_{\theta}}, \boldsymbol{W_{\phi}}, \boldsymbol{W_{g}} \in {\Bbb R}^{N \times C}$. 

As suggested in~\citep{wang2018non}, different choices for the function $f(.)$ are possible, including Gaussian and embedded Gaussian functions. The dot product similarity is most commonly used and is computed as, 
\begin{equation}
f(\theta(\boldsymbol{X}), \phi(\boldsymbol{X})) = \textit{softmax}\bigg ( \frac{\theta(\boldsymbol{X})\phi(\boldsymbol{X})^{T}}{\sqrt{d}}\bigg ),
\end{equation}
where $d$ corresponds to the dimension of the features. The non-local operation is finally wrapped into an attention block that is defined as:
\begin{equation}
    \begin{split}
        \boldsymbol{\hat{X}} = f(\theta(\boldsymbol{X}), \phi(\boldsymbol{X}))g(\boldsymbol{X}) + \boldsymbol{X}.
    \end{split}
\end{equation} 

\subsection{Nested block self-attention}
In the nested block self-attention, we split the feature map into spatially $M$ distinct memory blocks, $\textbf{X} = \{\textbf{x}_1, \ldots, \textbf{x}_m \}$. The non-local operation is done within each memory blocks by using a set of queries $\textbf{Q} = \{ \textbf{q}_1, \ldots, \textbf{q}_M\}$, keys $\textbf{K} = \{ \textbf{k}_1, \ldots, \textbf{k}_M\}$, and values $\textbf{V} = \{ \textbf{v}_1, \ldots, \textbf{v}_M\}$ corresponding to the block. A multi-headed attention computation is also possible by maintaining distinct weights for the $\textbf{Q,K,V}$ parameters. We however, chose to use shared weights for memory efficiency. But because we use overlapping memory blocks, the non-local attention itself is calculated as an aggregate of the attention from all blocks enclosing a pixel. Thus the attention $\beta_{j}$ for a feature $y_{j}$ at position $j$ is computed as a sum of non-local attentions from $P \leq M$ memory blocks it is contained in:
\begin{equation}
\label{eqn:NBSA}
\beta_j = \sum\limits_{p \in P}\textit{softmax}\bigg ( \frac{\theta^{p}(f_{j})\phi^{p}(f_j)^{T}}{\sqrt{d}}\bigg )g^{p}(f_j).
\end{equation}
\\
The first attention layer causes the attention information to flow from right to left and top to bottom as the image is scanned in raster scan order. Once the non-local attentions are aggregated for all the pixels, the non-local attention computed using the second NBSA layer naturally incorporates the attention flow in the reverse direction. This results in a bi-directional attention flow throughout the image and automatically extracts both the local and long-range dependencies. In our implementation, we separated the weights computed for the two attention blocks in order to increase the diversity in the attention calculation. An alternative memory efficient approach could also make use of a recurrent attention blocks, which we did not consider in this work. 

\subsection{Relative attention}
As a network design choice, we evaluated whether learning relative position embeddings improves segmentation accuracy as has previously been shown for object classification \citep{ramachandran2019standaloneSA} and sentence translations\citep{shaw2018self}. We trained the network to learn the relative attention weights or relative position embedding $E$ between pairs of features using an approach developed for sentence translations \citep{shaw2018self}. The relative attention weights are learned by representing the feature positions as nodes in a fully connected graph and modulating both the query-element similarity and the relative position weight of query-element as:
\begin{equation}
    \beta_j = \bigg (\frac{\theta(f_j)\phi(f_j)^{T} + \theta(f_j)E_j}{\sqrt{d}} \bigg )g(f_j).
\end{equation}
Memory efficient computation was achieved by using the \textit{Skew \/}\rm trick introduced in\citep{huang2018music}, wherein the relative position embedding is combined by padding the leftmost column with a dummy column vector of the same length as the number of positions, followed by reshaping of the matrix $QE^{T}$ to $(B+1,B)$, and then reslicing the matrix to retain the last \textit{B} rows and all columns to produce a $B \times B$ matrix.  

\subsection{Computational complexity}
The memory and computationally intensive non-local computations are restricted to the memory blocks of spatial extent $m = B \times B$. In the case of fully non-overlapping memory blocks, the whole computation of the various blocks can be parallelized across GPUs, thereby effectively requiring $\mathcal{O}(B^2)$ for the whole image. However, to enable the attention to flow between all the blocks, we use overlapping memory blocks, which requires to serialize the attention calculation, albeit the non-local computations still only require $\mathcal{O}(B^2)$. Given a stride $s, s \leq B$ used for placing the blocks, the computation for the entire image is now reduced from $\mathcal{O}(N^2)$ (when performing full non-local self attention) to $\mathcal{O}\Big(\Big(\frac{N}{B} + \frac{B}{s}\Big)^2\Big)$. 

\begin{figure*}
\centering
\includegraphics[width=0.9\textwidth,scale=0.1]{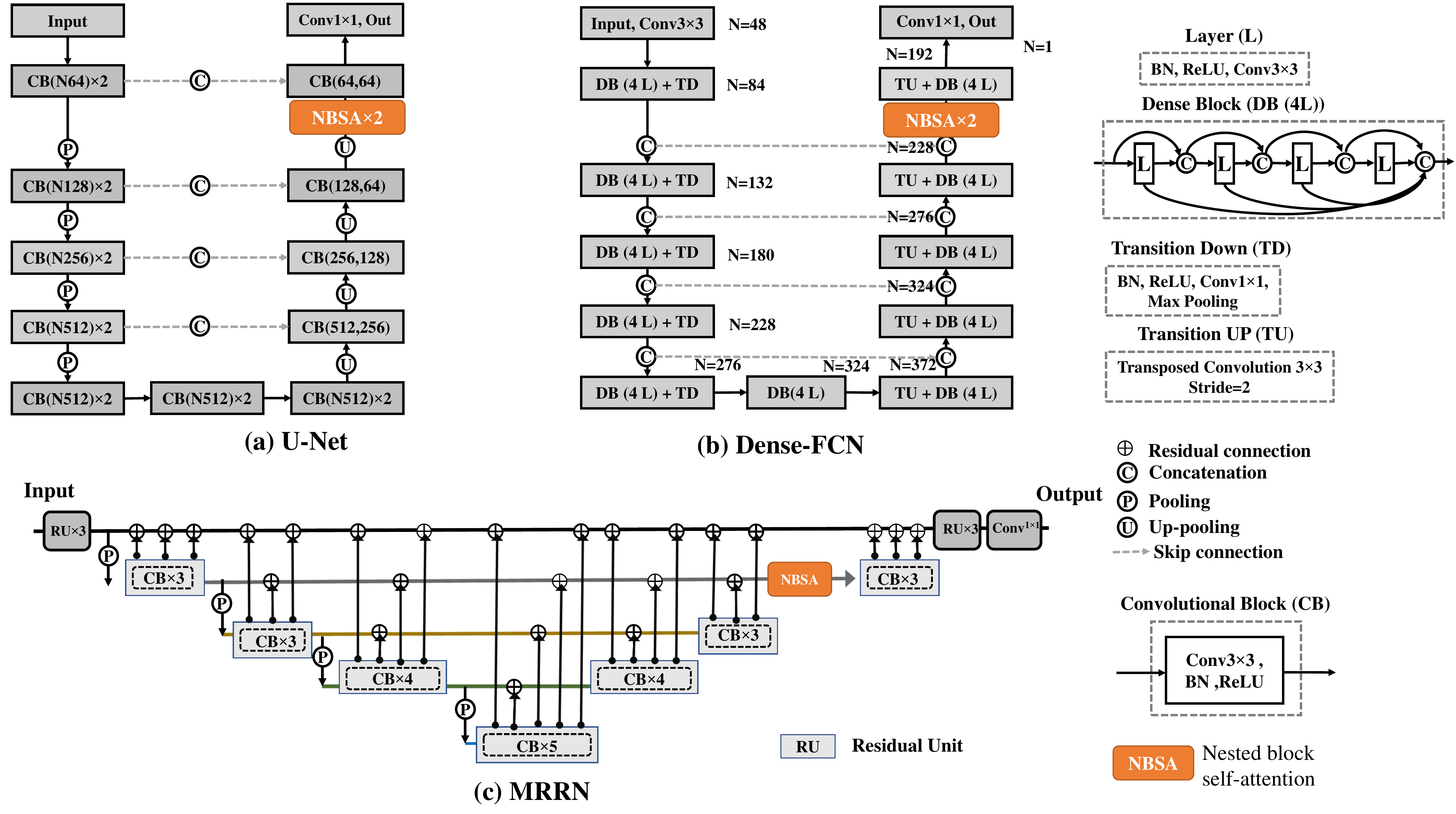}
\caption{\small {The Unet, DenseFCN, and MRRN networks used to implement the NBSA method.}} \label{fig:seg_structure}
\end{figure*}
\subsection{Implementation and Network Structure}

All networks were implemented using the Pytorch library ~\citep{paszke2017automatic} and  trained on Nvidia GTX 1080Ti with 12GB memory. The ADAM algorithm ~\citep{kingma2014adam} with an initial learning rate of 2e-4 was used during training. Training was performed by cropping the images to the head area and using image patches resized to 256 $\times$ 256, resulting in a total of 8000 training images. Models were constructed to segment multiple organs that included parotid glands, submandibular glands, brain stem, spinal cord, oral cavity, and manidble. 
\\
All methods were trained from scratch using the same training dataset with reasonable hyperparameter optimization done for equitable comparisons. The default kernel bandwidth size for NBSA was 36, resulting in a patch of size $36\times36$, a scanning stride of 24. The NBSA block was implemented in the penultimate layer of the various networks. Each layer consisted of a set of computational filters including convolutions (CONV), batch normalization (BN), and ReLU activations. 
%Architecture specific details of NBSA placement are provided in the Supplementary document. Thirty two feature channels were used in the attention blocks. 

\subsubsection{Unet with NBSA}
Our default network using the NBSA attention used a 2D U-net\citep{ronneberger2015u}. We modified U-net to include batch normalization after each convolution layer to speed up convergence. This network was composed of a series of convolutional blocks, with each block consisting of convolution, batch normalization, and ReLU activation. Skip connections are implemented to concatenate high-level and lower level features. Four max-pooling and four up-pooling layers were used to appropriately down- and up-sample features. The NBSA blocks were implemented in the penultimate layer with a feature size of 256$\times$256$\times$64. This network had 13.39 M parameters and 33 layers\footnote{layers are only counted on layers that have tunable weights} of Unet. 
\subsubsection{Dense fully convolutional network with NBSA}
The NBSA blocks were implemented in the penultimate layer of DenseFCN67\citep{jegou2017one} with a feature size of 128$\times$128$\times$228. This network is composed of Dense Blocks (DB)\citep{huang2017densely}, which successively concatenates feature maps computed from previous layers and increases the size of the feature maps. DB is produced by iterative concatenation of feature maps from previous layers within a block. A layer is composed of BN, ReLU, and 3$\times$3 convolution operation. Transition Down (TD) and Transition UP (TU) are used for down-sampling and up-sampling the feature size, respectively, where TD is composed of Batch Normalization, ReLU, 1$\times$1 convolution, 2$\times$2 max-pooling while TU is composed of 3$\times$3 transposed convolution. The DB used 4 feature concatenation layers, 5 TD, and 5 TU with a growing rate of 12. This resulted in 3.30 M parameters and 106 layers of DenseFCN.\\

\subsubsection{Multiple resolution residual network with NBSA}
Multiple resolution residual network (MRRN) \citep{jiangTMI2019} is a very deep network that we developed for generating segmentation of large and small lung tumors. This network combines aspects of both densely connected\citep{huang2017densely} and residual network\citep{HeResidualNet2016} architectures. Similar to the densely connected networks, features computed at multiple image resolutions are combined with the features in down-stream layers to provide higher image resolution information. Similar to residual networks, residual connections are used to increase the network's training stability as the network depth is increased.
\\
The MRRN network is composed of multiple residual feature streams, which carry feature maps at specific image resolution. There are as many residual feature streams as the number of pooling operations.  
\\
The main architectural component of the MRRN is the residual connection unit block (RCUB), which is composed of one or more serially connected residual connection units (RCU). RCU takes two inputs, feature map from the immediately preceding network layer or the output of preceding RCU and the feature map from the residual feature stream. Residual feature stream maps are processed starting from the nearest lower image resolution to the original image resolution by the successive RCUs. Each RCU consists of 3$\times$3 convolutions, BN, and
ReLU activation. The RCU has two outputs: one to connect to it's successive RCU and a second feature map that is passed back to the feature stream after appropriate upsampling through residual connections. The network is designed to detect both large and small structures in the image. Four max-pooling and up-pooling are used in the encoder and decoder in MRRN. The NBSA block is placed after the output of the layer preceding the last RCUB layer resulting in a feature size of 256$\times$256$\times$32. The MRRN resulted in 38.92M parameters.   

\subsection{Implementation of Unet-NBSA in routine clinical radiation treatment planning}

\begin{figure}
\centering
\includegraphics[width=0.5\textwidth,scale=0.1]{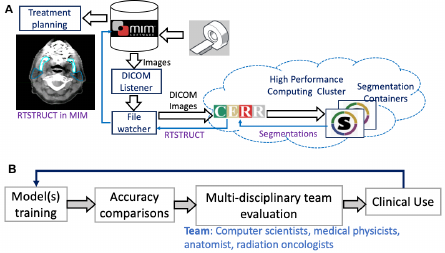}
\caption{\small {(A) Clinical implementation of the NBSA-Unet for radiation therapy planning. (B) Our approach for clinical software development and implementation.}}
\label{fig:clinicalworkflow}
\end{figure}

The Unet NBSA method is fully integrated into the clinical treatment planning workflow at our institution. Our software infrastructure framework called the ensemble of voxelwise attributes (EVA) directly interfaces with the commercial MIM software (MIM Software Inc., Beachwood, OH) used in the radiation therapy clinic, which features easy to use contour reviewing and editing functionalities (Figure~\ref{fig:clinicalworkflow} A). The EVA pipeline runs entirely in the background without any need for clinician intervention, thereby providing a seamless functionality without increasing clinician workload. DICOM listeners and file watcher clients automatically route DICOM images from specific anatomical site to appropriate deep learning method and place the segmented DICOM-RT structures with the associated DICOMs. The EVA pipeline supports batch processing and simultaneously manages multiple segmentation models for different disease sites. The segmentation models are encapsulated in operating system independent singularity containers and run on high-performance compute clusters. The container models can be swapped with no workflow disruption  for routine updates and performance improvements. The EVA pipeline is supported by the open-source CERR\citep{deasy2003cerr}, which performs DICOM handling, image processing, and converts segmentation masks into DICOM-RT structures. 
\\
We use a cyclical developmental cycle (Figure.~\ref{fig:clinicalworkflow} B) in order to adapt the methods to changes in imaging technology and incorporate clinical requirements. Clinically deployed deep learning methods are continuously improved and reviewed with a multi-disciplinary team of computer scientists, research and clinical physicists, radiation oncologists, and anatomists who meet on a weekly basis to qualitatively review individual cases as well as discuss future clinical needs. Staged evaluation using pre-clinical and periodic random clinical assessments are done to ensure robust ongoing performance.

\section{Experimental analysis} 
\begin{figure}
\centering
\includegraphics[width=0.5\textwidth,scale=0.3]{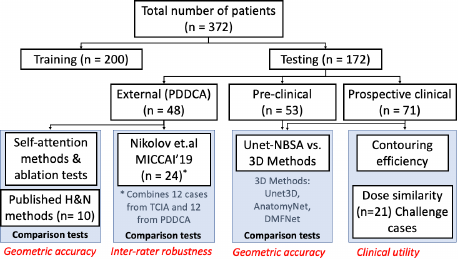}
\caption{\small {Flow-diagram of the training and testing done.} } \label{fig:REMARK}
\end{figure}

Figure.~\ref{fig:REMARK} summarizes the analyzed number of cases and experimental evaluations. Independent test sets from both the external public domain database for computational anatomy (PDDCA)\citep{raudaschl2017evaluation} and internal archive datasets were used. The PDDCA dataset was used to benchmark proposed method's performance and accuracy against  other self-attention methods and published HN segmentation methods. An internal institution dataset was used to evaluate contouring accuracy according to institutional preferences as well as with 3D architectures such as the 3DUnet\citep{nikolov2018deep}, the AnatomyNet\citep{zhu2018anatomynet}, and deep multi-fiber network (DFMNet)\citep{ChenDMFNet2019}. Both DFMNet and AnatomyNet methods used author provided implementations. Unet-NBSA is used in clinical treatment planning. Hence, the clinical metrics including contour adjustment and dosimetric metrics derived from the dose volume histogram (DVH) were also computed for this method. DVH analysis was performed on a random set of 21 cases containing high dose regions adjacent to the OARs in order to derive a worse case estimate of the DVH accuracies. 

\subsection{Performance metrics}
Geometric accuracy was computed by comparing with expert delineations using the Dice Similarity Coefficient (DSC), surface Dice similarity coefficient (SDSC), and Hausdorff metric at 95$^{th}$ percentile. Contour editing effort was summarized as contour adjustment ratio (CAR) and was computed as the ratio of the added path length (APL) to the total path length (TPL). APL\citep{vassenAPL2020} is the count of the number of boundary pixels added or deleted in the manual editing of the algorithm segmented contour. TPL is the total boundary pixels of the structure and corresponds to the effort needed in manual segmentation.
\\
Dosimetric metrics were derived from the dose volume histogram (DVH) computed from the algorithm and manual segmentations. These metrics consisted of absolute dose difference (ADD) of the mean and max doses, and the absolute difference in Vx measures (\% volume receiving at least xGy dose) for $x=5Gy, x=30Gy$, referred as $\Delta V5$ and $\Delta V30$, respectively. Spearman rank correlation coefficients were computed to compare the dose distributions summarized as mean dose. 

\section{Experimental results}
\begin{figure}
\centering
\includegraphics[width=0.5\textwidth,scale=0.1]{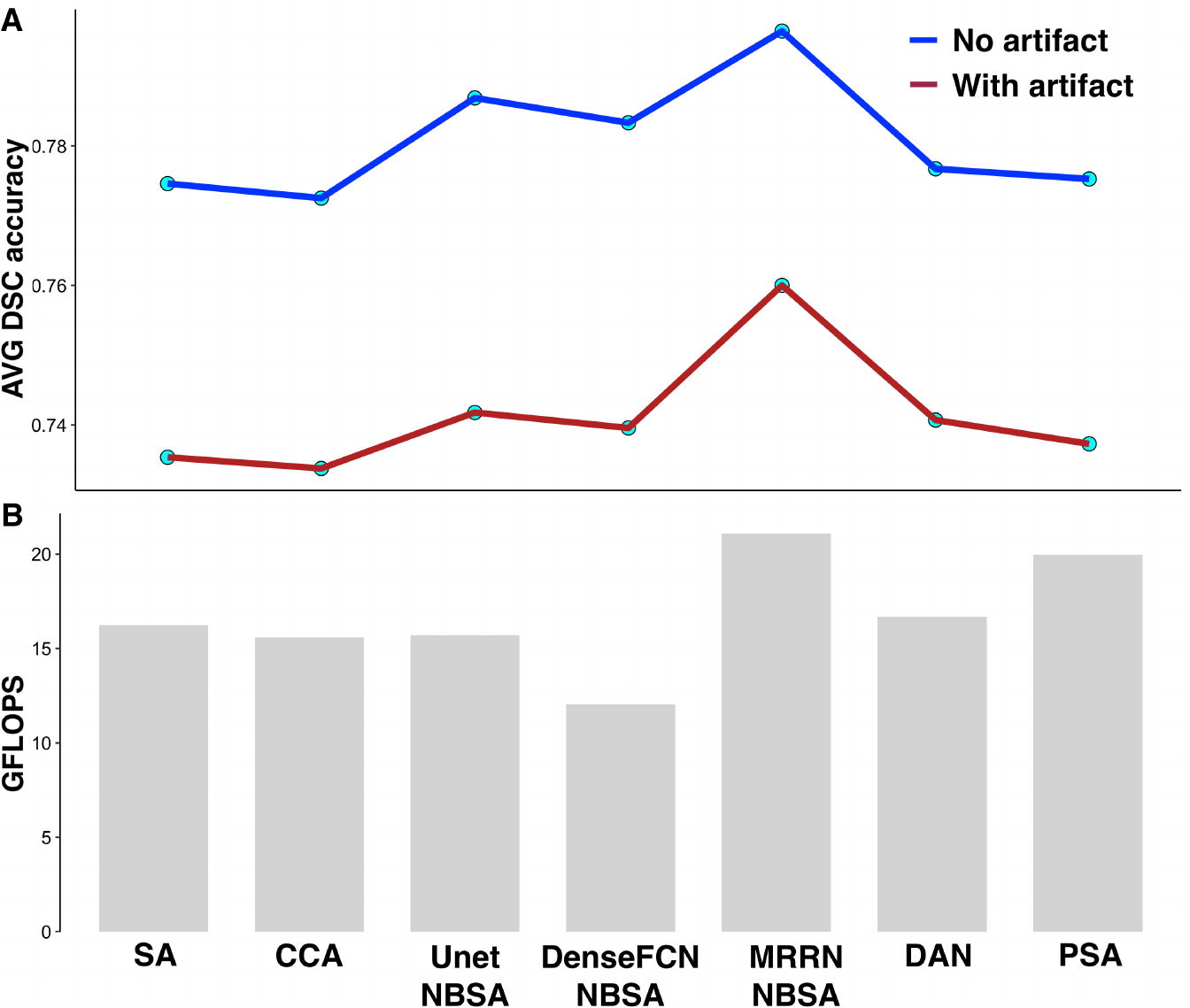}
\caption{\small {(A) Accuracy and (B) computational performance of the various self-attention networks.}} \label{fig:SelfAttentionComparison}
\end{figure}
\subsection{Computational performance and accuracy}
We compared the performance of the NBSA method against standard self-attention (SA)\citep{wang2018non}, the criss-cross self-attention (CCA)\citep{huang2019ccnet}, dual attention network (DAN)\citep{fu2018dual}, and the point-wise spatial attention (PSA)\citep{zhao2018psanet}. Open-source implementations of these methods provided by the authors were used. Unet network was used for all methods with the attention modules placed on the penultimate layer for all other methods. All methods were trained from scratch. As shown in Table.~\ref{tab:SA_comparisons}, the NBSA method was similarly computationally efficient but produced significantly more accurate brain stem and parotid glands segmentation than the CCA\citep{huang2019ccnet}.  
\\
Figure.~\ref{fig:SelfAttentionComparison} shows the average DSC accuracy and computational requirements of the various methods including, MRRN-NBSA and DenseFCN-NBSA on the PDDCA dataset. We distinguished performance in cases (n = 24) containing artifacts (artifacts present in at least 3 or more slices) and the rest (n=24). PDDCA cases did not contain any cases with resected organs and two cases had a tumor adjacent to submandibular glands. The NBSA methods were more accurate regardless of artifacts (Figure.~\ref{fig:SelfAttentionComparison} (A)). DenseFCN-NBSA was the most computationally efficient (12.05 GFLOPS), and produced significantly more accurate segmentations than CCA for the brain stem (DSC of 0.88, p $=$ 0.002) and the parotid glands (DSC of 0.842, p $=$ 0.037). MRRN NBSA required the most computations (21.11 GFLOPS), but it was significantly more accurate than CCA for brain stem (DSC of 0.89, p $< 0.001$), parotid glands (DSC of 0.86, p $< 0.001$), and submandibular glands (DSC of 0.80, p $=$ 0.01). 
\begin{table}
\caption{\label{tab:SA_comparisons} Geometric accuracy comparisons between the various self-attention methods and the NBSA method implemented on the Unet architecture. Accuracies that are significantly lower than Unet-NBSA are indicated with $^{*}$ for $<$ 0.1, $^{**}$ for $<$ 0.05, and $^{***}$ for  $<$ 0.01. BS - Brain stem, PG - Parotid glands, SG - submandibular glands, Man - mandible.} 
    \centering{
    \begin{tabular}{|c|c|c|c|c|c|} \hline
         \small{Method} & \small{BS} & \small{PG} & \small{SG} &  \small{Man} & \small{GFLOPS}\\ \hline
         \small{SA} & 0.87$^{***}$ & 0.84$^{*}$ & 0.78 & 0.89$^{***}$ & 16.24\\
         \small{CCA} & 0.86$^{***}$ & 0.84$^{**}$ & 0.77 & 0.92 & 15.60 \\
         \small{PSA} & 0.87$^{**}$ & 0.84 & 0.77 & 0.92 & 19.97 \\ 
         \small{Dual} & 0.88$^{*}$ & 0.85 & 0.76 & 0.92 & 16.69 \\
         \small{NBSA} & 0.89 & 0.86 & 0.79 & 0.93 & 15.71 \\ \hline
    \end{tabular}
    
    }
\end{table}

\begin{figure*}
\centering
\includegraphics[width=0.8\textwidth, scale=0.1]{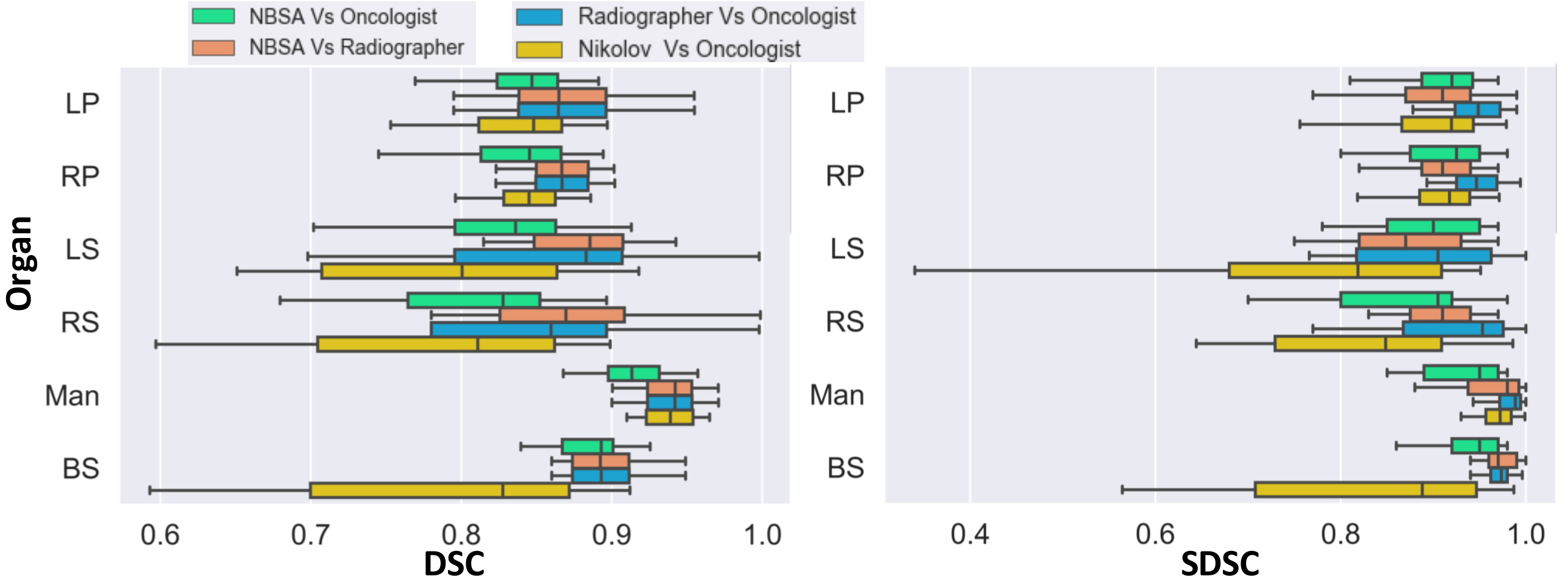}
\caption{\small {Organ specific performance measured using DSC and SDSC values computed using the radiation oncologist, radiographer as reference segmentation. Results provided by Nikolov et.al are also provided for comparison. }} 
\label{fig:seg_box_Nikolov}
\end{figure*}
\subsection{Comparison against published methods}
We next evaluated our method with respect to two-rater (radiographer and radiation oncologist) segmentations obtained from 24 HN cases provided by Nikolov et.al\citep{nikolov2018deep}. As shown in Figure.~\ref{fig:seg_box_Nikolov}, our method was as good or better than Nikolov et.al\citep{nikolov2018deep} with lower variability, particularly for the more difficult to segment submandibular glands and brain stem. Its variability was lower than the two raters for submandibular glands and similar to the two raters for parotid glands and brain stem. 
\\
Table~\ref{tab:Acc_PDDCA_test} shows the average DSC accuracy on the phase III PDDCA (n=16) cases  compared to multiple published methods that reported the performance on these same cases. Seven out of 16 cases had artifacts with 6 being the most number of slices with artifacts. There were two cases (\textit{0522c0746, 0522c0878\/}) \rm with an enlarged tumor abutting the ipsilateral submandibular gland. The MRRN-NBSA produced an average OAR DSC of 0.78 for both of these cases, while the Unet-NBSA produced an average DSC of 0.74 and 0.78, respectively. The CCA method on the other hand had a lower average DSC of 0.72 and 0.73. Nikolov et.al\citep{nikolov2018deep}, the only method to report case by case results reported a much lower average DSC of 0.67 and 0.72, respectively\footnote{Average DSC values are reported in \% in the paper}. 
\\
Compared to the published methods, our method used the fewest training examples. It is also the only method to use a 2D implementation besides Fritscher et.al\citep{fritscher2016deep}. Finally, our method was trained and validated using our internal archive dataset and then tested with the PDDCA as an entirely blinded and independent dataset, even when there are clear contouring differences. For example, our training contours for mandible overextend at least a mm outwards. Prior methods have therefore reported results by incorporating the Phase I and Phase II PDDCA data into the training\citep{liang2020,zhu2018anatomynet,nikolov2018deep,wang2018hierarchical}. Only the method in\citep{liang2020}, which reported the accuracy on PDDCA using the model trained on PDDCA was slightly better than our method. Our method's accuracy was as good or better than other published methods.

\begin{table*}  
				\centering{\caption{DSC accuracies for HN organ segmentation from Phase III PDDCA dataset. LP - Left parotid gland, RP - Right parotid gland, LS - Left submandibular gland, RS - Right submandibular gland, Man - Mandible, BS - Brain stem, AVG - Average over organs} 
					\label{tab:Acc_PDDCA_test} 
					\centering
					%\scriptsize
					%\tiny
					%\scriptsize
					%\footnotesize
					\small
					%\normalsize
					%\large
					%\Large
					%\LARGE
					%\huge
					%\Huge		
					\begin{tabular}{|c|c|c|c|c|c|c|c|c|c|} 
						\hline 
						
						{  Method  }  & {  2D/3D  }  &{Train}& {  LP  }& {  RP  }&{LS}&{RS}&{Man}&{BS}&{AVG}\\
						\hline

						{MICCAI 2015\citep{raudaschl2017evaluation}}&{-}&{-}&{0.84}&{0.84}&{0.78}&{0.78}&{0.93}&{0.88}&{0.84}\\			
						\hline

					   	{Nikolov et.al \citep{nikolov2018deep}}&{3D}&{663}&{0.87}&{0.85}&{0.76}&{0.78}&{0.94}&{0.80}&{0.83}\\
						\hline				
                        {Wang et.al \citep{wang2018hierarchical}}&{3D}&{663}&{0.83}&{0.83}&{-}&{-}&{0.94}&{0.90}&{-}\\

						%\hline				
						%{H ̈ansc et al\citep{hansch2018evaluation}}&{3D}&{254}&{$0.88$}&{$0.88$}&{-}&{-}&{-}&{-}&{-}\\
						\hline				
                       	{Fritscher et. al \citep{fritscher2016deep}}&{2D}&{}&{0.81}&{0.81}&{0.65}&{0.65}&{0.93}&{0.89}&{0.79}\\
						\hline 
						{AnatomyNet\citep{zhu2018anatomynet}}&{3D}&{261}&{0.88}&{0.87}&{0.81}&{0.81}&{0.93}&{0.87}&{0.86}\\
						\hline 
						{Liang et.al\citep{liang2020}} & {3D} & {PDDCA} & {0.87} & {0.87} & {0.81} & {0.81} & {0.94} & {0.92} & {\textcolor{red}{0.87}} 
						\\ 
					
					\hline	%{Unet-DPA}&{2D}&{102$^*$}&{0.86}&{0.87}&{0.80}&{0.81}&{0.94}&{0.90}&{0.87}\\
						\hline 	
{DenseFCN-NBSA}&{2D}&{200}&{0.86}&{0.86}&{0.76}&{0.77}&{0.93}&{0.88}&{0.85}\\
						\hline 	
{Unet-NBSA}&{2D}&{200}&{0.86}&{0.86}&{0.80}&{0.79}&{0.94}&{0.89}&{0.86}\\										\hline 	
{MRRN-NBSA}&{2D}&{200}&{0.87}&{0.87}&{0.81}&{0.81}&{0.94}&{0.90}&{0.87}\\	
						\hline 	
	\end{tabular}} 
\end{table*}		

\subsection{Geometric accuracy on internal archive test sets}
\begin{figure*}
\centering
\includegraphics[width=1.0\textwidth,scale=0.3]{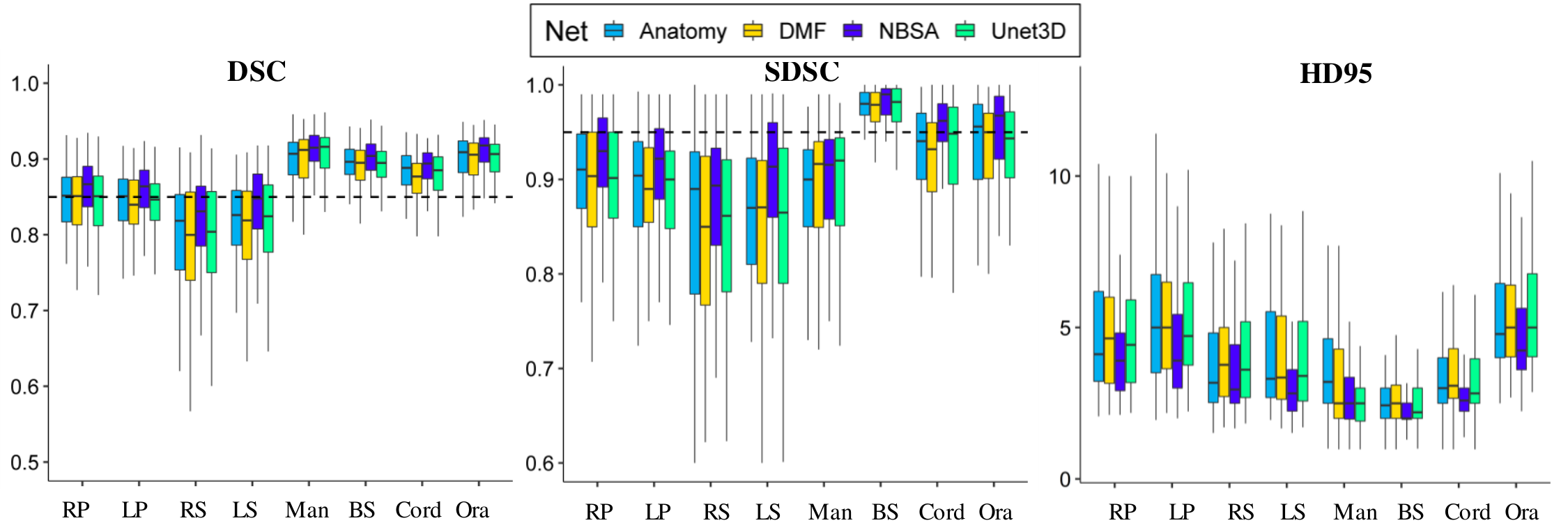}
\caption{\small {Box plots showing the accuracy of NBSA-Unet and multiple 3D deep network models on the internal archive dataset. The horizontal lines correspond to the mean inter-rater differences from Nikolov et.al\citep{nikolov2018deep}. LP - Left parotid, RP - Right parotid, LS - Left submandibular gland, RS - Right submandibular gland, Man - mandible, BS - Brain stem, Ora - Oral cavity.}} \label{fig:seg_box}
\end{figure*}

Figure.~\ref{fig:seg_box} shows the box plots of the geometric metrics computed for our and three different 3D network models on all of the internal testing dataset (n = 124). The reference lines in the plots correspond to the inter-rater difference derived between radiographer and a radiation oncologist from an entirely different dataset used in Nikolov et.al\citep{nikolov2018deep} and correspond to 0.85 for DSC and 0.935 for SDSC. The DSC accuracy of our method over all organs was 0.87 and the SDSC was 0.92. Our method also achieved a lower HD95 of 3.81mm compared with all other methods. The HD95 for all three methods were higher with DFMNet of 4.62mm, AnatomyNet of 4.48mm, and Unet3D of 4.18mm. The SDSC using our method exceeded 0.85 for all analyzed organs. The highest accuracy was achieved for brain stem with 0.97, followed by 0.95 for spinal cord, 0.92 for oral cavity, 0.92 for parotid gland, 0.91 for mandible, and 0.89 for submandibular glands. As shown in the subsequent section, these accuracies were sufficient to achieve high dosimetric accuracies. 

%compared with 0.86 for all the 3D methods. In particular, our method produced a DSC accuracy of 0.87 for parotid glands, 0.81 for submandibular glands, 0.90 for oral cavity, brain stem, and spinal cord, and 0.92 for mandible segmentation. AnatomyNet\citep{zhu2018anatomynet} was the next best accurate method, and achieved an accuracy of 0.85 for parotids, 0.80 for submandibular glands, 0.89 for brain stem, cord, and oral cavity, and 0.90 for the mandible. 

\begin{figure*}
\centering
\includegraphics[width=1\textwidth,scale=0.3]{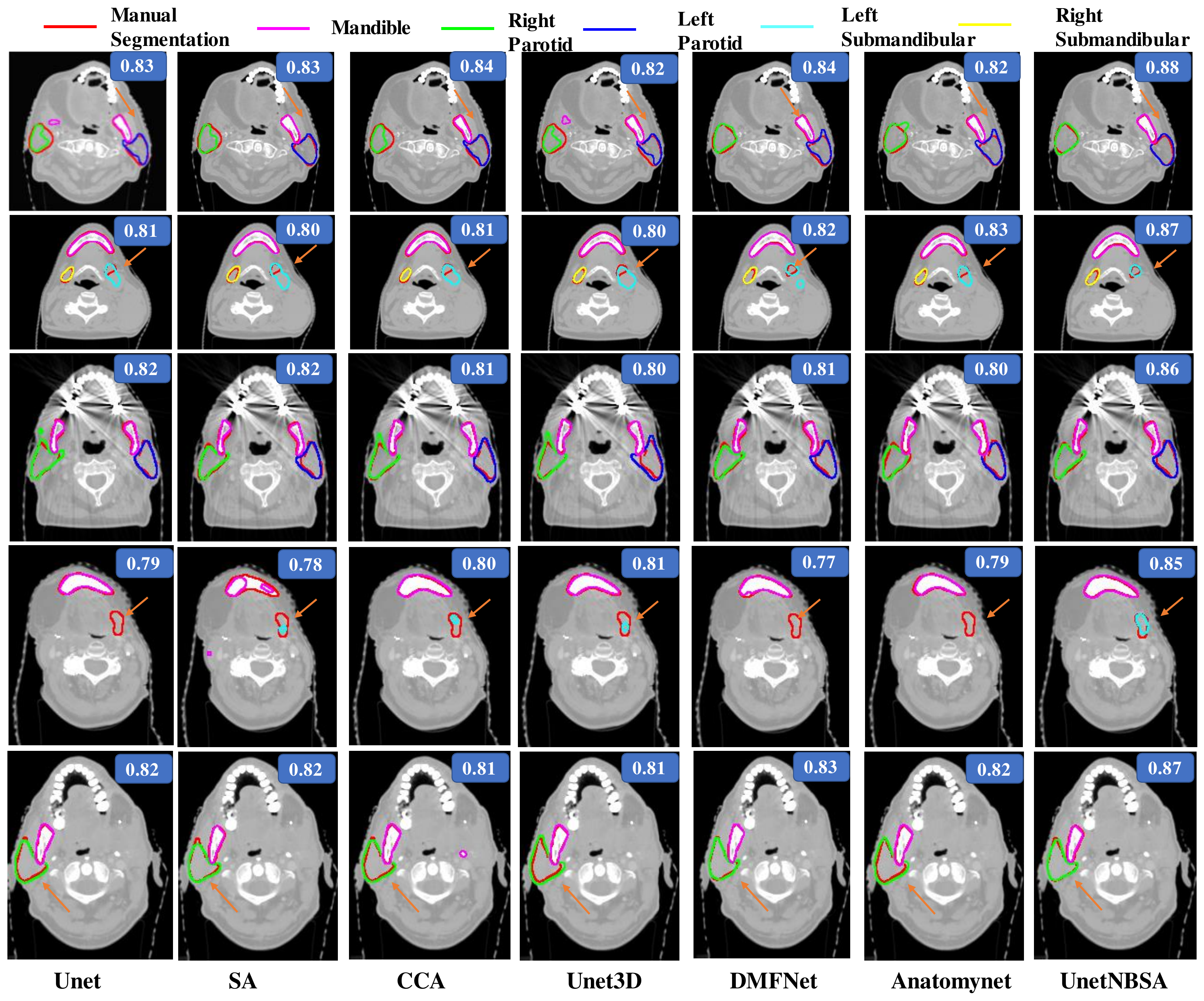}
\caption{\small {Segmentation results produced using our and multiple methods. Expert contours are in red while individual organs are depicted in unique colors. Arrows indicate the regions having highly variable segmentations. The overall DSC computed for all organs on this patient is shown in the top right corner of images.}} \label{fig:seg_overlay}
\end{figure*}

Figure.~\ref{fig:seg_overlay} shows some example clinical cases with large tumors abutting the normal organs or images with severe dental artifacts. As shown, our method produced a reasonably concordant with expert segmentation despite the difficult imaging conditions. We also show segmentations produced by two additional self-attention methods, namely the standard self-attention and the CCA method to illustrate the inability of these methods to handle the difficult imaging and complex anatomy. Similarly, other methods including 2D and 3DUnet, AnatomyNet, and DFMNet all produced worse segmentations for organs abutting tumors, where accuracy is crucial. 

\subsection{Clinical utility}
\begin{figure*}
\centering
\includegraphics[width=0.8\textwidth, scale=0.2]{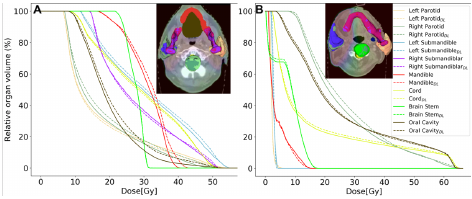}
\caption{\small {Comparison of dose volume histograms computed on the expert and deep learning segmentations with the high dose target region in (A) center and (B) right side of head and neck. Overlaid figures depict the algorithm and expert volumes, as well as the treatment dose distribution.}} \label{fig:dvh}
\end{figure*}

\begin{figure}
\centering
\includegraphics[width=0.5\textwidth, scale = 0.05]{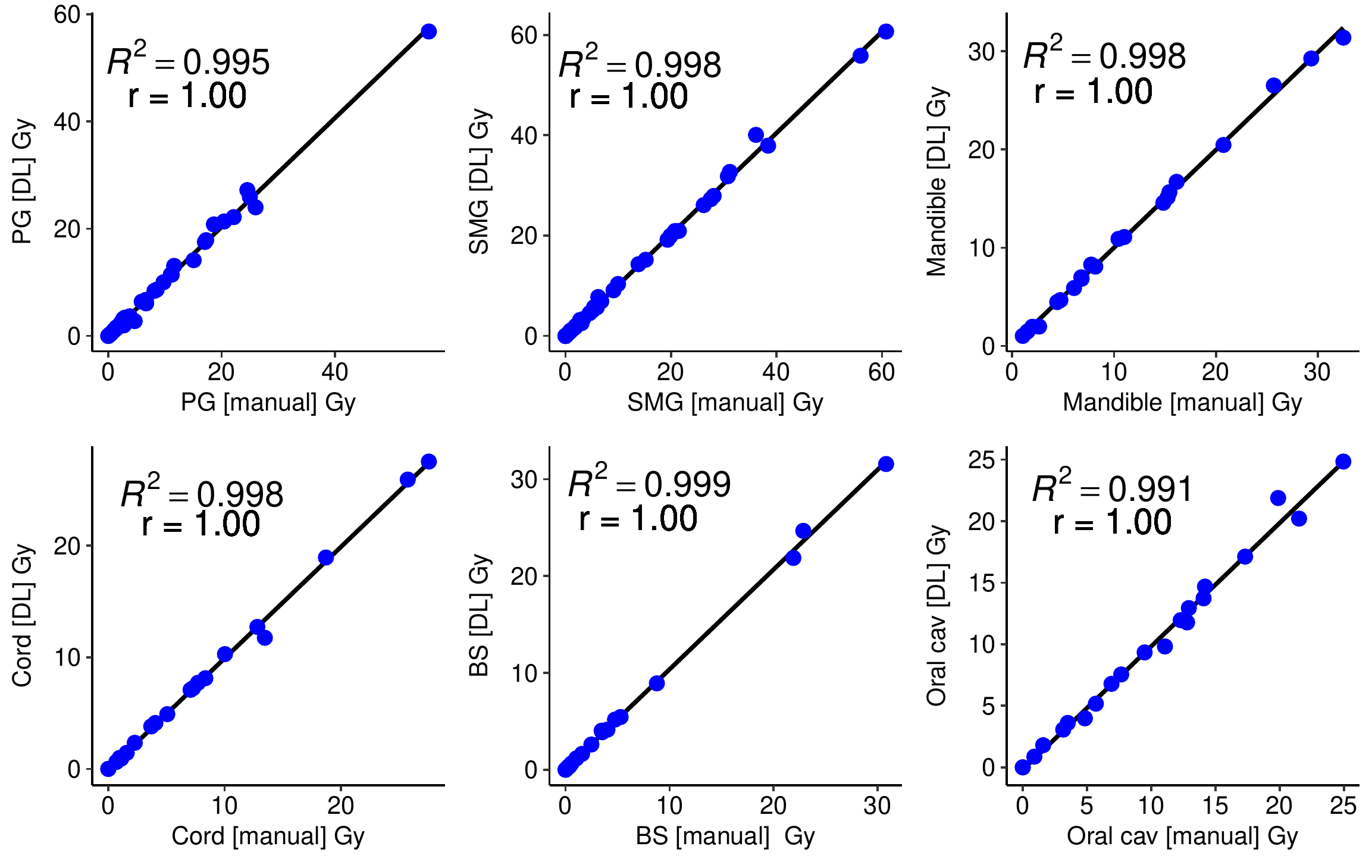}
\caption{\small {Scatter plots showing the mean dose computed using the DVH of the manual and deep learning (DL) segmentations. PG - Parotid glands, SMG - Submandiblar gland}} \label{fig:meandose}
\end{figure}

Table.~\ref{tab:ClinicalMetrics} summarizes the clinically relevant contour editing and dosimetric metrics. The spinal cord required only 16\% editing while oral cavity needed the most with 49\% editing, possibly due to the variation in the manual contouring. Overall, the CAR across all organs was 39\%, which corresponded to roughly 61\% reduction in contouring effort. Also important, the dose deviation between the algorithm and manual segmentations were within tolerable limits with the dose differences staying within 2Gy and below 2\% for all analyzed organs. 
\\
\begin{table}  
	\centering{\caption{Clinical utility performance metrics. CAR - Contour adjustment ratio, ADD - Absolute Dose Difference; RP - Right parotid, LP - Left parotid, RS - Right submandibular gland, LS - Left submandibular gland.} 
	\label{tab:ClinicalMetrics} 
	\centering
	\footnotesize
	\begin{tabular}{|c|c|c|c|c|c|}
	\hline 
	{Organ} & {CAR}& {max} & {mean} &{$\| \Delta V5\|$}&{$\| \Delta 30\|$} \\
	& \% & {ADD(Gy)} & {ADD(Gy)} & \% & \% \\
	\hline		
	{RP }&{46}&{0.47} & {0.60} & {0.84} & {0.84} \\
	{LP} & {45}&{0.53} & {0.34} & {1.66} & {1.65} \\ {RS} & {45} & {0.29} & {0.36} & {1.57} & {1.57}\\
	{LS}&{43} & {0.29} & {0.28} & {0.61} & {0.61}\\
	{Man}&{41} & {0.35}& {0.28} & {1.15} & {1.15}\\
	{Cord}&{16} & {0.83} & {0.17} & {1.21} & {1.21}\\
	{BS}&{30}&{0.58} & {0.23} & {1.06} & {1.07}\\
	{Oral} & {49} & {0.41} & {0.44} & {1.41} & {1.41} \\ \hline
    \end{tabular}} 
\end{table}		
Figure.~\ref{fig:dvh} shows the DVH computed for two different cases using both NBSA deep learning (DL) and manual segmentations. The corresponding volumetric segmentations are also depicted on the overlaid figures. Manual segmentations are shown in darker shades (dark blue for right parotid, orange for left parotid, purple for mandible, dark green for cord). Both of the selected cases contain the high dose target region, shown using a red colorwash adjacent to the organs.  Yet, the differences in the dose metrics derived using manual and deep learning (DL) segmentations were tolerable (that is dose difference below 2Gy). For instance,  for the organs shown in Figure.~\ref{fig:dvh} (A) only mandible with a mean ADD of 0.18Gy, $\| \Delta V30\|$ of 1.71\%, and the oral cavity with a mean ADD of 0.12Gy and $\| \Delta V30\|$ of 1.39\%, respectively had the highest dose differences among all other segmented OARs. Similarly, in Figure.~\ref{fig:dvh} (B), mandible (mean ADD of 0.20Gy, $\| \Delta V30\|$ of 1.89\%), cord (mean ADD of 0.17Gy, $\| \Delta V30\|$ of 2.98\%), and oral cavity (mean ADD of 1.24Gy, $\| \Delta V30\|$ of 3.91\%) had the highest dose differences, but all within tolerable limits.  
\\
Finally, the mean dose distributions computed from the algorithm and manual segmentations were strongly correlated for all organs (Figure.~\ref{fig:meandose}). 
\subsection{Network Design Experiments}
Network design tests were conducted to study network performance by varying: (i) the number of attention layers (single, dual [default], and three), (ii) placement of attention layers in the penultimate vs. last convolutional layer (a layer is a processing block composed of a sequence of convolutional layer (CONV), batch normalization (BN), and ReLU activation), (iii) memory block sizes (B=24, 36, 48), and (iv) overlapping \textcolor{black}{($s$=$B$$\times$$\frac{2}{3}$)} vs. non-overlapping \textcolor{black}{($s$=$B$)} memory blocks, where,  s is the stride length for placing the memory blocks in an image. We also evaluated the use of relative attention \citep{shaw2018self} for improving accuracy. 
\\
Our results showed that the dual attention layer configuration, which produces the bi-directional attention achieved sufficient accuracy without increasing computational needs as required when placing a third attention layer (Table.~\ref{tbl:ablation} (a)). 
\\
Similarly, the placement of self-attention in the penulatimate vs. last layer did not alter accuracy. 
On the other hand, placing the self attention in the last layer led to substantial increase (per 2D image) in the computational resources, clearly indicating that for the considered organs, self-attention placement on the penultimate layer was sufficient (Table.~\ref{tbl:ablation}(a,b)). 
\\
Table.~\ref{tbl:ablation}(c,d) shows the performance and computational times when using overlapping (default) and non-overlapping memory blocks. Our default implementation uses overlapping memory blocks in order to allow non-local information to pass through the whole image via the memory blocks. Our results indicate that non-overlapping memory blocks reduces the computational requirements and also the accuracy compared to the overlapping memory block configuration. 
\\
Increasing the memory block sizes did not contribute to accuracy. The analysis for memory block size of 48, with the self-attention placed on last layer was infeasible due to memory limitations and is not shown (Table.~\ref{tbl:ablation}(d)).
\\
Finally, the use of relative attention did not lead to accuracy differences. The mean DSC accuracies with and without relative attention were highly similar for the various organs (0.84 vs. 0.85 for parotid glands; 0.87 vs. 0.89 for brain stem, 0.79 each for submandibular glands, and 0.93 each for mandible).

\begin{table*}[tbp]
\caption{\small{Ablation tests to assess segmentation accuracy and computation time. } }
\centering
\begin{subtable}[t]{0.48\linewidth}
\centering
\vspace{0pt}
%\scriptsize
		%\tiny
		%\scriptsize
		\footnotesize
		%\small
		%\normalsize
		%\large
		%\Large
		%\LARGE
		%\huge
		%\Huge
\caption{\small{Number (N) of attention blocks placed on the \textbf{penultimate layer}.}}		
\begin{tabular}{|c|c|c|c|c|c|c|c|} 

			\hline 
			{  N   }  & {  LP  }  & {  RP }& {  LS  }& {  RS  }& {  M }&{  BS  }&{secs}\\
			\hline 
			\multirow{2}{*}{1}  &  {  0.83  }  & {  0.84 }& {  0.78  }& {  0.76  }& {  0.92 }&{ 0.87  }&{0.09}\\
			
            {  }  & {  0.05  }  & {  0.04 }& {  0.06  }& {  0.05  }& {  0.01 }&{ 0.02  }&{}\\			
			\hline 
			\multirow{2}{*}{2}  &  {  \textbf{0.85}  }  & {  \textbf{0.86} }& {  \textbf{0.79}  }& {  0.77  }& {  \textbf{0.93} }&{ \textbf{0.88}  }&{0.10}\\
			
            {  }  & {  0.04  }  & {  0.04 }& {  0.07  }& {  0.05  }& {  0.01 }&{ 0.02  }&{}\\	
            \hline
		\multirow{2}{*}{3}  &  {  0.85  }  & {  0.86 }& {  0.79  }& {  \textbf{0.78}  }& {  0.93 }&{ 0.88  }&{0.15}\\
			
            {  }  & {  0.05  }  & {  0.05 }& {  0.08  }& {  0.05  }& {  0.01 }&{ 0.02  }&{}\\	
            \hline
            
	\end{tabular} 
\label{tbl:iterNum_P}

\end{subtable}\hfill
\begin{subtable}[t]{0.48\linewidth}
\centering
\vspace{0pt}
%\scriptsize
		%\tiny
		%\scriptsize
		\footnotesize
		%\small
		%\normalsize
		%\large
		%\Large
		%\LARGE
		%\huge
		%\Huge
\caption{\small{Number (N) of attention blocks on \textbf{last layer}.}}		
\begin{tabular}{|c|c|c|c|c|c|c|c|} 

			\hline 
			{  N   }  & {  LP  }  & {  RP }& {  LS  }& {  RS  }& {  M }&{  BS  }&{secs}\\
			\hline 
			\multirow{2}{*}{1}  &  {  0.85  }  & {  0.85 }& {  0.78  }& {  0.75  }& {  0.92 }&{ 0.88  }&{0.24}\\
			
            {  }  & {  0.05  }  & {  0.04 }& {  0.08  }& {  0.09  }& {  0.02 }&{ 0.02  }&{}\\			
			\hline 
			\multirow{2}{*}{2}  &  {\textbf{0.85}}  & {\textbf{0.86} }& { \textbf{0.79} }& {  \textbf{0.79} }& { \textbf{ 0.93} }&{ \textbf{0.88}}&{0.44}\\
			
            {  }  & {  0.04  }  & {  0.04 }& {  0.07  }& {  0.06  }& {  0.02 }&{ 0.02  }&{}\\	
            \hline
		\multirow{2}{*}{3}  &  { 0.85  }  & {  0.86 }& {  0.79  }& {  0.79  }& {  0.93 }&{ 0.88  }&{0.65}\\
			
            {  }  & {  0.04  }  & {  0.04 }& {  0.06  }& {  0.05  }& {  0.02 }&{ 0.02  }&{}\\	
            \hline
	\end{tabular} 
\label{tbl:iterNum_L}

\end{subtable}\hfill

\begin{subtable}[t]{0.48\linewidth}
\centering
\vspace{0pt}
%\scriptsize
		%\tiny
		%\scriptsize
		\footnotesize
		%\small
		%\normalsize
		%\large
		%\Large
		%\LARGE
		%\huge
		%\Huge
\caption{\small{\textbf{Overlapping} blocks with varying sizes (B) in \textbf{penultimate layer}.}}		
\begin{tabular}{|c|c|c|c|c|c|c|c|} 

			\hline 
			\multicolumn{8}{|c|}{Overlapping attention blocks} \\
			\hline
			{B}  & {  LP  }  & {  RP }& {  LS  }& {  RS  }& {  M }&{  BS  }&{secs}\\
			\hline 
			
			\multirow{2}{*}{24}  &  {  0.84  }  & {  0.85 }& {  0.79  }& {  0.77  }& {  0.93 }&{ 0.88  }&{0.23}\\
			
            {  }  & {  0.06  }  & {  0.05 }& {  0.07  }& {  0.06  }& {  0.02 }&{ 0.02  }&{}\\			
			\hline 			
					\multirow{2}{*}{36}  &  {  \textbf{0.85}  }  & { \textbf{0.86} }& {  \textbf{0.79}  }& {  \textbf{0.77}  }& {  \textbf{0.93} }&{ \textbf{0.88}  }&{0.10}\\
			
            {  }  & {  0.04  }  & {  0.04 }& {  0.07  }& {  0.05  }& {  0.01 }&{ 0.02  }&{}\\	
			\hline
			\multirow{2}{*}{48}  &  {  0.85  }  & {  0.85 }& {  0.79  }& {  0.77  }& {  0.93 }&{ 0.88  }&{0.13}\\
			
            {  }  & {  0.05  }  & {  0.05 }& {  0.06  }& {  0.05  }& {  0.01 }&{ 0.01  }&{}\\			
			\hline 
			\hline 
			\multicolumn{8}{|c|}{Non-Overlapping attention blocks} \\
			\hline
			{B}  & {  LP  }  & {  RP }& {  LS  }& {  RS  }& {  M }&{  BS  }&{secs}\\
			\hline 
			
			\multirow{2}{*}{24}  &  {  0.84  }  & {  0.84 }& {  0.78  }& {  0.77  }& {  0.93 }&{ 0.88  }&{0.16}\\
			
            {  }  & {  0.05  }  & {  0.05 }& {  0.07  }& {  0.05  }& {  0.01 }&{ 0.02  }&{}\\			
			\hline 			
					\multirow{2}{*}{36}  &  {  \textbf{0.84}  }  & {  \textbf{0.85} }& {  \textbf{0.79}  }& {  \textbf{0.77}  }& {  \textbf{0.93} }&{ \textbf{0.88}  }&{0.09}\\
			
            {  }  & {  0.05  }  & {  0.04 }& {  0.07  }& {  0.06  }& {  0.01 }&{ 0.02  }&{}\\	
			\hline
			\multirow{2}{*}{48}  &  {  0.84  }  & {  0.85 }& {  0.79  }& {  0.77  }& {  0.93 }&{ 0.87  }&{0.09}\\
			
            {  }  & {  0.06  }  & {  0.06 }& {  0.06  }& {  0.06  }& {  0.01 }&{ 0.03  }&{}\\			
			\hline

			%\bottomrule
			
	\end{tabular} 
\label{tbl:over_P}

\end{subtable}\hfill
\begin{subtable}[t]{0.48\linewidth}
\centering
\vspace{0pt}
%\scriptsize
		%\tiny
		%\scriptsize
		\footnotesize
		%\small
		%\normalsize
		%\large
		%\Large
		%\LARGE
		%\huge
		%\Huge
		\caption{\small{\textbf{Overlapping} blocks with varying sizes (B) in \textbf{last layer}.}}
\begin{tabular}{|c|c|c|c|c|c|c|c|} 
            \hline
            \multicolumn{8}{|c|}{Overlap} \\
			\hline 
			{B}  & {  LP  }  & {  RP }& {  LS  }& {  RS  }& {  M }&{  BS  }&{secs}\\
			\hline 
			
			\multirow{2}{*}{24}  &  {  0.85  }  & {  0.86 }& {  0.79  }& {  0.78  }& {  0.93 }&{ 0.88  }&{0.88}\\
			
            {  }  & {  0.04  }  & {  0.04 }& {  0.06  }& {  0.06  }& {  0.01 }&{ 0.02  }&{}\\			
			\hline 			
			\multirow{2}{*}{36}  &  {  \textbf{0.85}  }  & {  \textbf{0.86} }& {  \textbf{0.79}  }& { \textbf{0.79}  }& {  \textbf{0.93} }&{ \textbf{0.88}  }&{0.44}\\
			
            {  }  & {  0.04  }  & {  0.04 }& {  0.07  }& {  0.06  }& {  0.02 }&{ 0.02  }&{}\\			
			\hline 	
			\multirow{2}{*}{48}  &  { -  }  & {  - }& {  -  }& { -  }& {  - }&{ -  }&{-}\\
			
            {  }  & {  -  }  & {  - }& {  - }& {  -  }& { - }&{-  }&{-}\\			
			\hline 
			\hline 
			\multicolumn{8}{|c|}{Non-Overlap} \\
			\hline
	{B}  & {  LP  }  & {  RP }& {  LS  }& {  RS  }& {  M }&{  BS  }&{secs}\\
			\hline 			
			\multirow{2}{*}{24}  &  {  0.85  }  & {  0.85 }& {  0.77  }& {  0.76  }& {  0.92 }&{ 0.88  }&{0.42}\\
			
            {  }  & {  0.05  }  & {  0.04 }& {  0.09  }& {  0.07  }& {  0.02 }&{ 0.02  }&{}\\			
			\hline 			
			\multirow{2}{*}{36}  &  {  \textbf{0.84}  }  & {  \textbf{0.85} }& {  \textbf{0.79}  }& {  \textbf{0.78}  }& {  \textbf{0.93} }&{ \textbf{0.88}  }&{0.23}\\
			
            {  }  & {  0.05  }  & {  0.05 }& {  0.06  }& {  0.06  }& {  0.01 }&{ 0.02  }&{}\\			
			\hline 	
			\multirow{2}{*}{48}  &  { -  }  & {  - }& {  -  }& { -  }& {  - }&{ -  }&{}\\
			
            {  }  & {  -  }  & {  - }& {  - }& {  -  }& { - }&{-  }&{}\\			
			\hline 				
			%\bottomrule
			
	\end{tabular} 
\label{tbl:over_L}

\end{subtable}\hfill

\label{tbl:ablation}
\end{table*}

\subsection{Qualitative results using attention map}

\begin{figure}
\centering
\includegraphics[width=0.48\textwidth]{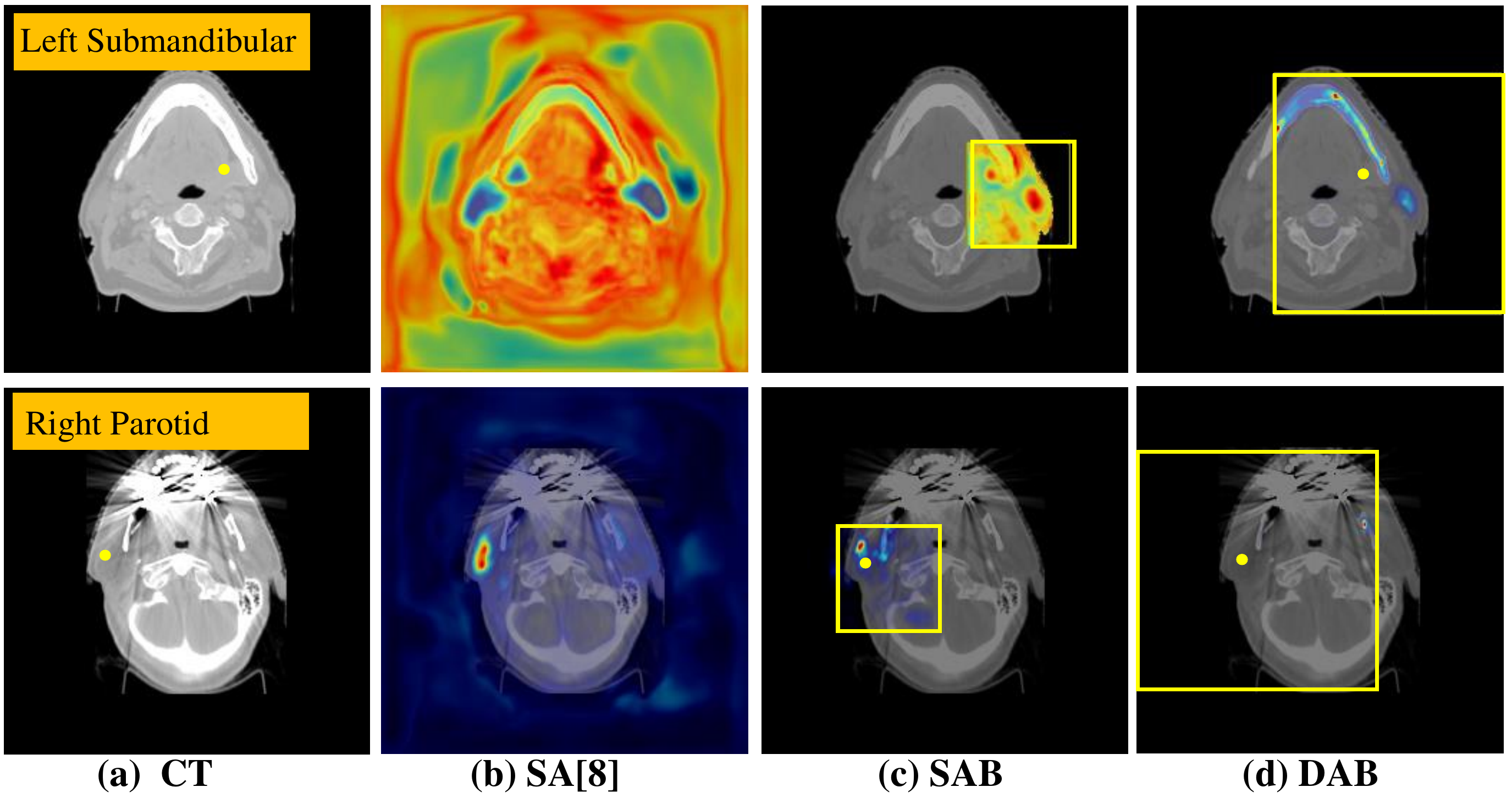}
\caption{\small Self attention maps from representative cases for interest point (shown in yellow) using (b) non-local self attention, and proposed approach using (c) single block attention and (d) nested block self-attention. Yellow rectangle indicates the effective contextual field. } \label{fig:attention map}
\end{figure}

Figure~\ref{fig:attention map} shows self-attention maps generated from representative CT example cases when using single self-attention block layer (SAB) and dual attention block (DAB) layer. The attention blocks were placed on the penultimate layer (our default configuration). The self-attention map produced using the non-local self-attention (SA) is also shown for comparison purposes. As expected, SA method results in activations in various parts of the image, some of which may not contain any useful information (e.g. background air). In comparison, our approach extracts the locally relevant information in the first attention layer (Figure.~\ref{fig:attention map}(c)) and relevant distant organ information in the second layer (Figure.~\ref{fig:attention map}(d)). The non-local activations are sparsified and only those regions that provide useful information for localization and segmentation are activated, which may contribute to higher robustness to abnormal anatomy and difficult imaging conditions. For example, while the SA method still contains some activations from the areas containing dental artifacts, none of those pixels are activated using our approach, effectively forcing the network to rely on the more reliable local information for segmentation. 

\section{Discussion}

In this work, we developed a new computationally efficient and robust approach using nested block self-attention for generating segmentations of multiple organs in the head and neck in difficult, yet routinely seen clinical cases. Performance evaluation was done using both open-source and internally collected retrospective and prospective clinical datasets, with a total of 172 analyzed cases. No exclusion criteria related to imaging conditions (artifacts or abnormal anatomy) was applied in the internal clinical datasets. Our method has been implemented for routine clinical treatment planning at our institution since May, 2020. To our best knowledge, this is the first report of a deep learning approach evaluated on routine clinical treatment planning cases.
\\
We performed a comprehensive  performance evaluation of our method using conventionally used Dice and Hausdorff metrics, as well as clinically relevant dosimetric and contouring effort metrics. We demonstrated feasibility of implementing the nested block self-attention method into multiple deep learning networks with varying number of parameters.
\\
Our method was as good or better than the 3D networks and achieved sufficient dosimetric accuracy for clinical use. Importantly, our method was robust to images with artifacts and abnormal anatomy such as abutting tumors or missing anatomy. Robust performance in the presence of severe dental artifacts and anatomical abnormalities is essential for routine use of automated segmentation method. This study is one of the first studies to specifically examine the performance under difficult and real clinical conditions. Majority of the evaluated deep learning methods make use of the open-source PDDCA\citep{raudaschl2017evaluation}, which has some artifacts and two cases with tumors abutting organs in the dataset. But the number of such difficult cases in the phase III, the testing dataset is low. The only other deep learning study to our knowledge to have done a large scale evaluation on clinical datasets, albeit using retrospective cases is the work by van Dijk et.al\citep{vandijk2020}. But this study had only 18\% cases with dental artifacts, which were subjected to artifact reduction and cases with salivary gland tumors were excluded. Consequently, although these methods have demonstrated high accuracies on specific datasets, they may not produce sufficient accuracy on common, yet difficult clinical conditions. For example, the accuracy of other state-of-the-art 3D methods was highly comparable to ours on the Phase III PDDCA dataset. But their performance, such as the 3DUnet and AnatomyNet\citep{zhu2018anatomynet} decreased especially on the difficult cases in the clinical dataset. 
\\
In this study, we chose to not retrain our method on the PDDCA training set for evaluation on the PDDCA testing set as done by several others\citep{liang2020,zhu2018anatomynet,nikolov2018deep,wang2018hierarchical}. This is because, we wanted to ascertain the performance on a blinded independent dataset despite known contouring differences. For instance, our clinical segmentations slightly "over-segment" the mandible compared to PDDCA cases in order to ensure maximal radiation dose sparing. Such a blinded evaluation is essential to understand the generalization performance, especially of deep learning models due to the large number of model parameters. We found that our method more closely approximated the inter-rater segmentations, especially for small organs with low soft-tissue contrast such as the submandibular glands compared to the Unet3D method\citep{nikolov2018deep}, the only method to publish case-by-case analysis accuracy for side-by-side comparisons. In order to allow reproducible comparisons, we also provide our performance metrics for these same cases in the Supplementary document.  
\\
Clinical evaluation of our method also showed that the dosimetric performance was highly correlated to the corresponding metrics extracted using manual delineations. This indicates that this method can produce reliable clinical performance. Also important, our method resulted in an overall contour adjustment of 39\% or conversely resulted in a saving of contour adjustment for 61\% of the segmented contours. 
\\
Finally, the nested block self-attention method is generally applicable to different network structures and resulted in as low computational requirements as the CCA\citep{huang2019ccnet} while achieving better accuracy, especially in such cases with dental artifacts and abnormal anatomy. All of these results indicate that our method was robust to commonly seen difficult clinical conditions and feasible as an automated segmentation tool for clinical radiation treatment planning. 
\\
Our approach has a few limitations, such as lack of segmentation of structures such as the optic structures, because segmentations for these organs were not available in all of our training sets. Ours is also currently implemented as a 2D segmentation method. A 3D self-attention method is planned future work. However, the performance of our method was same or better than 3D implementations including the AnatomyNet\citep{zhu2018anatomynet} and 3DUnet methods. Finally, the choice of the attention block is currently a tunable parameter that depends on the application, specifically the size of the image used for the analysis. 

\section{Conclusion}
We developed a new computationally efficient nested block self-attention segmentation method that is robust to clinically difficult conditions such as dental artifacts and abnormal anatomy seen on routine radiation therapy datasets. This is the first report of a clinically implemented deep learning technique for head and neck cancer radiation therapy. We also performed comprehensive evaluation of our method against multiple methods and using multiple performance metrics on a large dataset and demonstrated feasibility of our approach for clinical use.  
\section*{Acknowledgments}
\textcolor{black}{This work was supported by the MSK Cancer Center support grant/core grant
P30 CA008748.}
%\section*{References}
		%%Harvard
%		\bibliographystyle{model2-names.bst}
%		\biboptions{authoryear}
%		\bibliography{mybibliography}

\bibliographystyle{model2-names}\biboptions{authoryear}
\bibliography{ref}

\begin{thebibliography}{49}
\expandafter\ifx\csname natexlab\endcsname\relax\def\natexlab#1{#1}\fi
\providecommand{\url}[1]{\texttt{#1}}
\providecommand{\href}[2]{#2}
\providecommand{\path}[1]{#1}
\providecommand{\DOIprefix}{doi:}
\providecommand{\ArXivprefix}{arXiv:}
\providecommand{\URLprefix}{URL: }
\providecommand{\Pubmedprefix}{pmid:}
\providecommand{\doi}[1]{\href{http://dx.doi.org/#1}{\path{#1}}}
\providecommand{\Pubmed}[1]{\href{pmid:#1}{\path{#1}}}
\providecommand{\bibinfo}[2]{#2}
\ifx\xfnm\relax \def\xfnm[#1]{\unskip,\space#1}\fi
%Type = Inproceedings
\bibitem[{Chen et~al.(2019)Chen, Liu, Ding, Zheng and Li}]{ChenDMFNet2019}
\bibinfo{author}{Chen, C.}, \bibinfo{author}{Liu, X.}, \bibinfo{author}{Ding,
  M.}, \bibinfo{author}{Zheng, J.}, \bibinfo{author}{Li, J.},
  \bibinfo{year}{2019}.
\newblock \bibinfo{title}{3\textsc{D} dilated multi-fiber network for real-time
  brain tumor segmentation in \textsc{MRI}}, in: \bibinfo{editor}{Shen, D.},
  \bibinfo{editor}{Liu, T.}, \bibinfo{editor}{Peters, T.M.},
  \bibinfo{editor}{Staib, L.H.}, \bibinfo{editor}{Essert, C.},
  \bibinfo{editor}{Zhou, S.}, \bibinfo{editor}{Yap, P.T.},
  \bibinfo{editor}{Khan, A.} (Eds.), \bibinfo{booktitle}{Medical Image
  Computing and Computer Assisted Intervention -- MICCAI 2019},
  \bibinfo{publisher}{Springer}. pp. \bibinfo{pages}{184--192}.
%Type = Article
\bibitem[{Chen et~al.()Chen, Huang, Lin, Qi, Xie, Wei, Chang, Sun, Wu and
  Lu}]{chen2020}
\bibinfo{author}{Chen, H.}, \bibinfo{author}{Huang, D.}, \bibinfo{author}{Lin,
  L.}, \bibinfo{author}{Qi, Z.}, \bibinfo{author}{Xie, P.},
  \bibinfo{author}{Wei, J.}, \bibinfo{author}{Chang, L.}, \bibinfo{author}{Sun,
  Y.}, \bibinfo{author}{Wu, D.}, \bibinfo{author}{Lu, Y.}, .
\newblock \bibinfo{title}{Prior attention enhanced convolutional neural network
  based automatic segmentation of organs at risk for head and neck cancer
  radiotherapy}.
\newblock \bibinfo{journal}{IEEE Access} \bibinfo{volume}{8},
  \bibinfo{pages}{179018--179027}.
\newblock \DOIprefix\doi{10.1109/ACCESS.2020.3028038}.
%Type = Inproceedings
\bibitem[{Chen et~al.(2018)Chen, Kalantidis, Yan and Feng}]{Chen2018A2NetsDA}
\bibinfo{author}{Chen, Y.}, \bibinfo{author}{Kalantidis, Y.},
  \bibinfo{author}{Yan, S.}, \bibinfo{author}{Feng, J.}, \bibinfo{year}{2018}.
\newblock \bibinfo{title}{A$^2$-nets: Double attention networks}, in:
  \bibinfo{booktitle}{Proc. Neural Information Processing Systems}.
%Type = Article
\bibitem[{Deasy et~al.(2003)Deasy, Blanco and Clark}]{deasy2003cerr}
\bibinfo{author}{Deasy, J.O.}, \bibinfo{author}{Blanco, A.I.},
  \bibinfo{author}{Clark, V.H.}, \bibinfo{year}{2003}.
\newblock \bibinfo{title}{\textsc{CERR}: a computational environment for
  radiotherapy research}.
\newblock \bibinfo{journal}{Med Phys} \bibinfo{volume}{30},
  \bibinfo{pages}{979--985}.
%Type = Article
\bibitem[{van Dijk et~al.(2020)van Dijk, den Bosch, Aljabar, Peressutti, Both,
  Steenbakkers, Langendijk, Gooding and Brouwer}]{vandijk2020}
\bibinfo{author}{van Dijk, L.}, \bibinfo{author}{den Bosch, L.},
  \bibinfo{author}{Aljabar, P.}, \bibinfo{author}{Peressutti, D.},
  \bibinfo{author}{Both, S.}, \bibinfo{author}{Steenbakkers, R.},
  \bibinfo{author}{Langendijk, J.}, \bibinfo{author}{Gooding, M.},
  \bibinfo{author}{Brouwer, C.}, \bibinfo{year}{2020}.
\newblock \bibinfo{title}{Improving automatic delineation for head and neck
  organs at risk by deep learning contouring}.
\newblock \bibinfo{journal}{Radiother Oncol} \bibinfo{volume}{142},
  \bibinfo{pages}{115--123}.
%Type = Inproceedings
\bibitem[{Fritscher et~al.(2016)Fritscher, Raudaschl, Zaffino, Spadea, Sharp
  and Schubert}]{fritscher2016deep}
\bibinfo{author}{Fritscher, K.}, \bibinfo{author}{Raudaschl, P.},
  \bibinfo{author}{Zaffino, P.}, \bibinfo{author}{Spadea, M.F.},
  \bibinfo{author}{Sharp, G.C.}, \bibinfo{author}{Schubert, R.},
  \bibinfo{year}{2016}.
\newblock \bibinfo{title}{Deep neural networks for fast segmentation of 3d
  medical images}, in: \bibinfo{booktitle}{International Conference on Medical
  Image Computing and Computer-Assisted Intervention},
  \bibinfo{organization}{Springer}. pp. \bibinfo{pages}{158--165}.
%Type = Article
\bibitem[{Fu et~al.(2018)Fu, Liu, Tian, Fang and Lu}]{fu2018dual}
\bibinfo{author}{Fu, J.}, \bibinfo{author}{Liu, J.}, \bibinfo{author}{Tian,
  H.}, \bibinfo{author}{Fang, Z.}, \bibinfo{author}{Lu, H.},
  \bibinfo{year}{2018}.
\newblock \bibinfo{title}{Dual attention network for scene segmentation}.
\newblock \bibinfo{journal}{arXiv preprint arXiv:1809.02983} .
%Type = Inproceedings
\bibitem[{{Gao} et~al.(2019){Gao}, {Huang}, {Chen}, {Wang}, {Deng}, {Chen},
  {Yang}, {Zhang}, {Tao} and {Li}}]{gao2019MICCAI}
\bibinfo{author}{{Gao}, Y.}, \bibinfo{author}{{Huang}, R.},
  \bibinfo{author}{{Chen}, M.}, \bibinfo{author}{{Wang}, Z.},
  \bibinfo{author}{{Deng}, J.}, \bibinfo{author}{{Chen}, Y.},
  \bibinfo{author}{{Yang}, Y.}, \bibinfo{author}{{Zhang}, J.},
  \bibinfo{author}{{Tao}, C.}, \bibinfo{author}{{Li}, H.},
  \bibinfo{year}{2019}.
\newblock \bibinfo{title}{{\textsc{F}ocus\textsc{N}et: \textsc{I}mbalanced
  Large and Small Organ Segmentation with an End-to-End Deep Neural Network for
  Head and Neck CT Images}}, in: \bibinfo{booktitle}{Medical Image Computing
  and Computer Assisted Intervention -- MICCAI 2019},
  \bibinfo{publisher}{Springer}, \bibinfo{address}{Cham}. pp.
  \bibinfo{pages}{829--838}.
%Type = Inproceedings
\bibitem[{{Guo} et~al.(2020){Guo}, {Jin}, {Zhu}, {Ho}, {Harrison}, {Chao},
  {Xiao} and {Lu}}]{guo2020}
\bibinfo{author}{{Guo}, D.}, \bibinfo{author}{{Jin}, D.},
  \bibinfo{author}{{Zhu}, Z.}, \bibinfo{author}{{Ho}, T.Y.},
  \bibinfo{author}{{Harrison}, A.P.}, \bibinfo{author}{{Chao}, C.H.},
  \bibinfo{author}{{Xiao}, J.}, \bibinfo{author}{{Lu}, L.},
  \bibinfo{year}{2020}.
\newblock \bibinfo{title}{Organ at risk segmentation for head and neck cancer
  using stratified learning and neural architecture search}, in:
  \bibinfo{booktitle}{2020 IEEE/CVF Conference on Computer Vision and Pattern
  Recognition (CVPR)}, pp. \bibinfo{pages}{4222--4231}.
\newblock \DOIprefix\doi{10.1109/CVPR42600.2020.00428}.
%Type = Article
\bibitem[{Haq et~al.(2019)Haq, Berry, Deasy, Hunt and Veeraraghavan}]{haq2019}
\bibinfo{author}{Haq, R.}, \bibinfo{author}{Berry, S.}, \bibinfo{author}{Deasy,
  J.}, \bibinfo{author}{Hunt, M.}, \bibinfo{author}{Veeraraghavan, H.},
  \bibinfo{year}{2019}.
\newblock \bibinfo{title}{Dynamic multiatlas selection-based consensus
  segmentation of head and neck structures from ct images}.
\newblock \bibinfo{journal}{Med Phys} \bibinfo{volume}{46},
  \bibinfo{pages}{5612--5622}.
%Type = Article
\bibitem[{Harari et~al.(2010)Harari, Song and Tome}]{Harari2010}
\bibinfo{author}{Harari, P.}, \bibinfo{author}{Song, S.},
  \bibinfo{author}{Tome, W.}, \bibinfo{year}{2010}.
\newblock \bibinfo{title}{Emphasizing conformal avoidance versus target
  definition for \textsc{IMRT} planning in head-and-neck cancer}.
\newblock \bibinfo{journal}{Int J Radiat Oncol Biol Phys} \bibinfo{volume}{77},
  \bibinfo{pages}{950--8}.
%Type = Inproceedings
\bibitem[{{He} et~al.(2016){He}, {Zhang}, {Ren} and {Sun}}]{HeResidualNet2016}
\bibinfo{author}{{He}, K.}, \bibinfo{author}{{Zhang}, X.},
  \bibinfo{author}{{Ren}, S.}, \bibinfo{author}{{Sun}, J.},
  \bibinfo{year}{2016}.
\newblock \bibinfo{title}{Deep residual learning for image recognition}, in:
  \bibinfo{booktitle}{2016 IEEE Conference on Computer Vision and Pattern
  Recognition (CVPR)}, pp. \bibinfo{pages}{770--778}.
\newblock \DOIprefix\doi{10.1109/CVPR.2016.90}.
%Type = Article
\bibitem[{Hoang-Duc et~al.()Hoang-Duc, Eminowicz, Mendes, Wong, McClelland,
  Modat, Cardoso, Mendelson, Veiga, Kadir and Ourselin}]{duc2015}
\bibinfo{author}{Hoang-Duc, A.}, \bibinfo{author}{Eminowicz, G.},
  \bibinfo{author}{Mendes, R.}, \bibinfo{author}{Wong, S.L.},
  \bibinfo{author}{McClelland, J.}, \bibinfo{author}{Modat, M.},
  \bibinfo{author}{Cardoso, M.}, \bibinfo{author}{Mendelson, A.},
  \bibinfo{author}{Veiga, C.}, \bibinfo{author}{Kadir, T.},
  \bibinfo{author}{Ourselin, S.}, .
\newblock \bibinfo{title}{Validation of clinical acceptability of an
  atlas-based segmentation algorithm for the delineation of organs at risk in
  head and neck cancer}.
\newblock \bibinfo{journal}{Med Phys} \bibinfo{volume}{42},
  \bibinfo{pages}{5027--34}.
%Type = Inproceedings
\bibitem[{Hu et~al.(2018)Hu, Gu, Zhang, Dai and Wei}]{hu2018relation}
\bibinfo{author}{Hu, H.}, \bibinfo{author}{Gu, J.}, \bibinfo{author}{Zhang,
  Z.}, \bibinfo{author}{Dai, J.}, \bibinfo{author}{Wei, Y.},
  \bibinfo{year}{2018}.
\newblock \bibinfo{title}{Relation networks for object detection}, in:
  \bibinfo{booktitle}{Proceedings of the IEEE Conference on Computer Vision and
  Pattern Recognition}, pp. \bibinfo{pages}{3588--3597}.
%Type = Article
\bibitem[{Hu et~al.(2020)Hu, Shen, Albanie, Sun and
  Wu}]{Hu2020SqueezeExcitation}
\bibinfo{author}{Hu, J.}, \bibinfo{author}{Shen, L.}, \bibinfo{author}{Albanie,
  S.}, \bibinfo{author}{Sun, G.}, \bibinfo{author}{Wu, E.},
  \bibinfo{year}{2020}.
\newblock \bibinfo{title}{Squeeze-and-excitation networks}.
\newblock \bibinfo{journal}{IEEE Trans. Pattern Anal. Mach. Intell.}
  \bibinfo{volume}{42}, \bibinfo{pages}{2011–2023}.
\newblock \DOIprefix\doi{10.1109/TPAMI.2019.2913372}.
%Type = Inproceedings
\bibitem[{Huang et~al.(2018)Huang, Vaswani, Uszkoreit, Simon, Hawthorne,
  Shazeer, Dai, Hoffman, Dinculescu and Eck}]{huang2018music}
\bibinfo{author}{Huang, C.Z.A.}, \bibinfo{author}{Vaswani, A.},
  \bibinfo{author}{Uszkoreit, J.}, \bibinfo{author}{Simon, I.},
  \bibinfo{author}{Hawthorne, C.}, \bibinfo{author}{Shazeer, N.},
  \bibinfo{author}{Dai, A.M.}, \bibinfo{author}{Hoffman, M.D.},
  \bibinfo{author}{Dinculescu, M.}, \bibinfo{author}{Eck, D.},
  \bibinfo{year}{2018}.
\newblock \bibinfo{title}{Music transformer}, in:
  \bibinfo{booktitle}{International conference on Learning Representation
  (ICLR)}.
%Type = Inproceedings
\bibitem[{Huang et~al.(2017)Huang, Liu, Van Der~Maaten and
  Weinberger}]{huang2017densely}
\bibinfo{author}{Huang, G.}, \bibinfo{author}{Liu, Z.}, \bibinfo{author}{Van
  Der~Maaten, L.}, \bibinfo{author}{Weinberger, K.Q.}, \bibinfo{year}{2017}.
\newblock \bibinfo{title}{Densely connected convolutional networks}, in:
  \bibinfo{booktitle}{Proceedings of the IEEE conference on computer vision and
  pattern recognition}, pp. \bibinfo{pages}{4700--4708}.
%Type = Article
\bibitem[{Huang et~al.(2019)Huang, Yuan, Guo, Zhang, Chen and
  Wang}]{Huang2019InterlacedSS}
\bibinfo{author}{Huang, L.}, \bibinfo{author}{Yuan, Y.}, \bibinfo{author}{Guo,
  J.}, \bibinfo{author}{Zhang, C.}, \bibinfo{author}{Chen, X.},
  \bibinfo{author}{Wang, J.}, \bibinfo{year}{2019}.
\newblock \bibinfo{title}{Interlaced sparse self-attention for semantic
  segmentation}.
\newblock \bibinfo{journal}{ArXiv} \bibinfo{volume}{abs/1907.12273}.
%Type = Inproceedings
\bibitem[{{Huang} et~al.(2019){Huang}, {Wang}, {Huang}, {Huang}, {Wei} and
  {Liu}}]{huang2019ccnet}
\bibinfo{author}{{Huang}, Z.}, \bibinfo{author}{{Wang}, X.},
  \bibinfo{author}{{Huang}, L.}, \bibinfo{author}{{Huang}, C.},
  \bibinfo{author}{{Wei}, Y.}, \bibinfo{author}{{Liu}, W.},
  \bibinfo{year}{2019}.
\newblock \bibinfo{title}{\textsc{CCNet}: Criss-cross attention for semantic
  segmentation}, in: \bibinfo{booktitle}{2019 {IEEE/CVF} International
  Conference on Computer Vision, {ICCV} 2019, Seoul, Korea (South), October 27
  - November 2, 2019}, \bibinfo{publisher}{{IEEE}}. pp.
  \bibinfo{pages}{603--612}.
\newblock \URLprefix \url{https://doi.org/10.1109/ICCV.2019.00069},
  \DOIprefix\doi{10.1109/ICCV.2019.00069}.
%Type = Article
\bibitem[{Ibragimov and Xing(2017)}]{ibragimov2017}
\bibinfo{author}{Ibragimov, B.}, \bibinfo{author}{Xing, L.},
  \bibinfo{year}{2017}.
\newblock \bibinfo{title}{Segmentation of organs-at-risk in head and neck ct
  images using convolutional neural networks}.
\newblock \bibinfo{journal}{Med Phys} \bibinfo{volume}{44},
  \bibinfo{pages}{547--555}.
%Type = Inproceedings
\bibitem[{J{\'e}gou et~al.(2017)J{\'e}gou, Drozdzal, Vazquez, Romero and
  Bengio}]{jegou2017one}
\bibinfo{author}{J{\'e}gou, S.}, \bibinfo{author}{Drozdzal, M.},
  \bibinfo{author}{Vazquez, D.}, \bibinfo{author}{Romero, A.},
  \bibinfo{author}{Bengio, Y.}, \bibinfo{year}{2017}.
\newblock \bibinfo{title}{The one hundred layers tiramisu: Fully convolutional
  densenets for semantic segmentation}, in: \bibinfo{booktitle}{Proceedings of
  the IEEE Conference on Computer Vision and Pattern Recognition Workshops},
  pp. \bibinfo{pages}{11--19}.
%Type = Article
\bibitem[{{Jiang} et~al.(2019){Jiang}, {Hu}, {Liu}, {Halpenny}, {Hellmann},
  {Deasy}, {Mageras} and {Veeraraghavan}}]{jiangTMI2019}
\bibinfo{author}{{Jiang}, J.}, \bibinfo{author}{{Hu}, Y.C.},
  \bibinfo{author}{{Liu}, C.J.}, \bibinfo{author}{{Halpenny}, D.},
  \bibinfo{author}{{Hellmann}, M.D.}, \bibinfo{author}{{Deasy}, J.O.},
  \bibinfo{author}{{Mageras}, G.}, \bibinfo{author}{{Veeraraghavan}, H.},
  \bibinfo{year}{2019}.
\newblock \bibinfo{title}{Multiple resolution residually connected feature
  streams for automatic lung tumor segmentation from ct images}.
\newblock \bibinfo{journal}{IEEE Transactions on Medical Imaging}
  \bibinfo{volume}{38}, \bibinfo{pages}{134--144}.
\newblock \DOIprefix\doi{10.1109/TMI.2018.2857800}.
%Type = Article
\bibitem[{Kingma and Ba(2014)}]{kingma2014adam}
\bibinfo{author}{Kingma, D.P.}, \bibinfo{author}{Ba, J.}, \bibinfo{year}{2014}.
\newblock \bibinfo{title}{\textsc{Adam}: A method for stochastic optimization}.
\newblock \bibinfo{journal}{Proceedings of the 3rd International Conference on
  Learning Representations (ICLR)} .
%Type = Article
\bibitem[{Liang et~al.(2020)Liang, Thung, Nie, Zhang and Shen}]{liang2020}
\bibinfo{author}{Liang, S.}, \bibinfo{author}{Thung, K.H.},
  \bibinfo{author}{Nie, D.}, \bibinfo{author}{Zhang, Y.},
  \bibinfo{author}{Shen, D.}, \bibinfo{year}{2020}.
\newblock \bibinfo{title}{Multi-view spatial aggregation framework for joint
  localization and segmentation of organs at risk in head and neck \textsc{CT}
  images}.
\newblock \bibinfo{journal}{IEEE Trans Med Imaging} \bibinfo{volume}{39},
  \bibinfo{pages}{2794--2805}.
%Type = Article
\bibitem[{Liu et~al.(2020)Liu, Lei, Fu, Wang, Zhou, Jiang, McDonald, Beitler,
  Curran, Liu and Yang}]{liu2020}
\bibinfo{author}{Liu, Y.}, \bibinfo{author}{Lei, Y.}, \bibinfo{author}{Fu, Y.},
  \bibinfo{author}{Wang, T.}, \bibinfo{author}{Zhou, J.},
  \bibinfo{author}{Jiang, X.}, \bibinfo{author}{McDonald, M.},
  \bibinfo{author}{Beitler, J.}, \bibinfo{author}{Curran, W.},
  \bibinfo{author}{Liu, T.}, \bibinfo{author}{Yang, X.}, \bibinfo{year}{2020}.
\newblock \bibinfo{title}{Head and neck multi-organ auto-segmentation on ct
  images aided by synthetic \textsc{MRI}}.
\newblock \bibinfo{journal}{Med Phys} \DOIprefix\doi{10.1002/mp.14378}.
%Type = Article
\bibitem[{Men et~al.()Men, Geng, Cheng, Zhong, Huang, Fan, Plastaras, Lin and
  Xiao}]{men2019}
\bibinfo{author}{Men, K.}, \bibinfo{author}{Geng, H.}, \bibinfo{author}{Cheng,
  C.}, \bibinfo{author}{Zhong, H.}, \bibinfo{author}{Huang, M.},
  \bibinfo{author}{Fan, Y.}, \bibinfo{author}{Plastaras, J.},
  \bibinfo{author}{Lin, A.}, \bibinfo{author}{Xiao, Y.}, .
\newblock \bibinfo{title}{Technical note: \textsc{M}ore accurate and efficient
  segmentation of organs-at-risk in radiotherapy with convolutional neural
  networks cascades}.
\newblock \bibinfo{journal}{Med Phys} \bibinfo{volume}{46},
  \bibinfo{pages}{286--292}.
\newblock \DOIprefix\doi{10.1002/mp.13296}.
%Type = Article
\bibitem[{Nelms et~al.(2010)Nelms, Tome, Robinson and Wheeler}]{nelms2010}
\bibinfo{author}{Nelms, B.}, \bibinfo{author}{Tome, W.},
  \bibinfo{author}{Robinson, G.}, \bibinfo{author}{Wheeler, J.},
  \bibinfo{year}{2010}.
\newblock \bibinfo{title}{Variations in the contouring of organs at risk: test
  case from a patient with oropharyngeal cancer}.
\newblock \bibinfo{journal}{Int J Radiat Oncol Phys} \bibinfo{volume}{82},
  \bibinfo{pages}{368--78}.
%Type = Article
\bibitem[{Nikolov et~al.(2018)Nikolov, Blackwell, Mendes, De~Fauw, Meyer,
  Hughes, Askham, Romera-Paredes, Karthikesalingam, Chu
  et~al.}]{nikolov2018deep}
\bibinfo{author}{Nikolov, S.}, \bibinfo{author}{Blackwell, S.},
  \bibinfo{author}{Mendes, R.}, \bibinfo{author}{De~Fauw, J.},
  \bibinfo{author}{Meyer, C.}, \bibinfo{author}{Hughes, C.},
  \bibinfo{author}{Askham, H.}, \bibinfo{author}{Romera-Paredes, B.},
  \bibinfo{author}{Karthikesalingam, A.}, \bibinfo{author}{Chu, C.}, et~al.,
  \bibinfo{year}{2018}.
\newblock \bibinfo{title}{Deep learning to achieve clinically applicable
  segmentation of head and neck anatomy for radiotherapy}.
\newblock \bibinfo{journal}{arXiv preprint arXiv:1809.04430} .
%Type = Inproceedings
\bibitem[{Oktay et~al.(2018)Oktay, Schlemper, Folgoc, Lee, Heinrich, Misawa,
  Mori, McDonagh, Hammerla, Kainz, Glocker and Rueckert}]{oktay2018}
\bibinfo{author}{Oktay, O.}, \bibinfo{author}{Schlemper, J.},
  \bibinfo{author}{Folgoc, J.}, \bibinfo{author}{Lee, M.},
  \bibinfo{author}{Heinrich, M.}, \bibinfo{author}{Misawa, K.},
  \bibinfo{author}{Mori, K.}, \bibinfo{author}{McDonagh, S.},
  \bibinfo{author}{Hammerla, N.}, \bibinfo{author}{Kainz, B.},
  \bibinfo{author}{Glocker, B.}, \bibinfo{author}{Rueckert, D.},
  \bibinfo{year}{2018}.
\newblock \bibinfo{title}{Attention \textsc{U}-\textsc{N}et: Learning where to
  look for the pancreas}, in: \bibinfo{booktitle}{Proc. Machine Learning in
  Medical Imaging}.
%Type = Inproceedings
\bibitem[{Parmar et~al.(2018)Parmar, Vaswani, Uszkoreit, Kaiser, Shazeer and
  Ku}]{parmar2018image}
\bibinfo{author}{Parmar, N.}, \bibinfo{author}{Vaswani, A.},
  \bibinfo{author}{Uszkoreit, J.}, \bibinfo{author}{Kaiser, {\L}.},
  \bibinfo{author}{Shazeer, N.}, \bibinfo{author}{Ku, A.},
  \bibinfo{year}{2018}.
\newblock \bibinfo{title}{Image transformer}, in: \bibinfo{editor}{Dy, J.},
  \bibinfo{editor}{Krause, A.} (Eds.), \bibinfo{booktitle}{Proceedings of the
  35th International Conference on Machine Learning},
  \bibinfo{publisher}{PMLR}, \bibinfo{address}{StockholmsmÃ€ssan, Stockholm
  Sweden}. pp. \bibinfo{pages}{4055--4064}.
%Type = Article
\bibitem[{Paszke et~al.(2017)Paszke, Gross, Chintala, Chanan, Yang, DeVito,
  Lin, Desmaison, Antiga and Lerer}]{paszke2017automatic}
\bibinfo{author}{Paszke, A.}, \bibinfo{author}{Gross, S.},
  \bibinfo{author}{Chintala, S.}, \bibinfo{author}{Chanan, G.},
  \bibinfo{author}{Yang, E.}, \bibinfo{author}{DeVito, Z.},
  \bibinfo{author}{Lin, Z.M.}, \bibinfo{author}{Desmaison, A.},
  \bibinfo{author}{Antiga, L.}, \bibinfo{author}{Lerer, A.},
  \bibinfo{year}{2017}.
\newblock \bibinfo{title}{Automatic differentiation in
  \textsc{P}y\textsc{T}orch} .
%Type = Article
\bibitem[{{Ramachandran} et~al.(2019){Ramachandran}, {Parmar}, {Vaswani},
  {Bello}, {Levskaya} and {Shlens}}]{ramachandran2019standaloneSA}
\bibinfo{author}{{Ramachandran}, P.}, \bibinfo{author}{{Parmar}, N.},
  \bibinfo{author}{{Vaswani}, A.}, \bibinfo{author}{{Bello}, I.},
  \bibinfo{author}{{Levskaya}, A.}, \bibinfo{author}{{Shlens}, J.},
  \bibinfo{year}{2019}.
\newblock \bibinfo{title}{{Stand-Alone Self-Attention in Vision Models}}.
\newblock \bibinfo{journal}{arXiv} , \bibinfo{pages}{arXiv:1906.05909}.
%Type = Article
\bibitem[{Raudaschl et~al.(2017)Raudaschl, Zaffino, Sharp, Spadea, Chen,
  Dawant, Albrecht, Gass, Langguth, L{\"u}thi et~al.}]{raudaschl2017evaluation}
\bibinfo{author}{Raudaschl, P.F.}, \bibinfo{author}{Zaffino, P.},
  \bibinfo{author}{Sharp, G.C.}, \bibinfo{author}{Spadea, M.F.},
  \bibinfo{author}{Chen, A.}, \bibinfo{author}{Dawant, B.M.},
  \bibinfo{author}{Albrecht, T.}, \bibinfo{author}{Gass, T.},
  \bibinfo{author}{Langguth, C.}, \bibinfo{author}{L{\"u}thi, M.}, et~al.,
  \bibinfo{year}{2017}.
\newblock \bibinfo{title}{Evaluation of segmentation methods on head and neck
  ct: Auto-segmentation challenge 2015}.
\newblock \bibinfo{journal}{Med Phys} \bibinfo{volume}{44},
  \bibinfo{pages}{2020--2036}.
%Type = Article
\bibitem[{Ren et~al.()Ren, Xiang, Nie, Shao, Zhang, Shen and Wang}]{ren18}
\bibinfo{author}{Ren, X.}, \bibinfo{author}{Xiang, L.}, \bibinfo{author}{Nie,
  D.}, \bibinfo{author}{Shao, Y.}, \bibinfo{author}{Zhang, H.},
  \bibinfo{author}{Shen, D.}, \bibinfo{author}{Wang, Q.}, .
\newblock \bibinfo{title}{Interleaved 3\textsc{D-CNN}s for joint segmentation
  of small-volume structures in head and neck \textsc{CT} images}.
\newblock \bibinfo{journal}{Med Phys} \bibinfo{volume}{45},
  \bibinfo{pages}{2063--2075}.
\newblock \DOIprefix\doi{10.1002/mp.12837}.
%Type = Inproceedings
\bibitem[{Ronneberger et~al.(2015)Ronneberger, Fischer and
  Brox}]{ronneberger2015u}
\bibinfo{author}{Ronneberger, O.}, \bibinfo{author}{Fischer, P.},
  \bibinfo{author}{Brox, T.}, \bibinfo{year}{2015}.
\newblock \bibinfo{title}{U-net: Convolutional networks for biomedical image
  segmentation}, in: \bibinfo{booktitle}{International Conference on Medical
  Image Computing and Computer-Assisted Intervention (MICCAI)},
  \bibinfo{organization}{Springer}. pp. \bibinfo{pages}{234--241}.
%Type = Inproceedings
\bibitem[{Roy et~al.(2018)Roy, Navab and Wachinger}]{roy2018concurrent}
\bibinfo{author}{Roy, A.G.}, \bibinfo{author}{Navab, N.},
  \bibinfo{author}{Wachinger, C.}, \bibinfo{year}{2018}.
\newblock \bibinfo{title}{Concurrent spatial and channel ‘squeeze \&
  excitation’in fully convolutional networks}, in:
  \bibinfo{booktitle}{International Conference on Medical Image Computing and
  Computer-Assisted Intervention}, \bibinfo{organization}{Springer}. pp.
  \bibinfo{pages}{421--429}.
%Type = Inproceedings
\bibitem[{Shaw et~al.(2018)Shaw, Uszkoreit and Vaswani}]{shaw2018self}
\bibinfo{author}{Shaw, P.}, \bibinfo{author}{Uszkoreit, J.},
  \bibinfo{author}{Vaswani, A.}, \bibinfo{year}{2018}.
\newblock \bibinfo{title}{Self-attention with relative position
  representations}, in: \bibinfo{booktitle}{Proceedings of the 2018 Conference
  of the North {A}merican Chapter of the Association for Computational
  Linguistics: Human Language Technologies, Volume 2 (Short Papers)},
  \bibinfo{publisher}{Association for Computational Linguistics},
  \bibinfo{address}{New Orleans, Louisiana}. pp. \bibinfo{pages}{464--468}.
\newblock \DOIprefix\doi{10.18653/v1/N18-2074}.
%Type = Article
\bibitem[{Shen et~al.(2018)Shen, Zhou, Long, Jiang and Zhang}]{shen2018bi}
\bibinfo{author}{Shen, T.}, \bibinfo{author}{Zhou, T.}, \bibinfo{author}{Long,
  G.}, \bibinfo{author}{Jiang, J.}, \bibinfo{author}{Zhang, C.},
  \bibinfo{year}{2018}.
\newblock \bibinfo{title}{Bi-directional block self-attention for fast and
  memory-efficient sequence modeling}.
\newblock \bibinfo{journal}{arXiv preprint arXiv:1804.00857} .
%Type = Article
\bibitem[{{Tang} et~al.(2019){Tang}, {Tang}, {Xiao} and
  {Summers}}]{Tang2019CCNet}
\bibinfo{author}{{Tang}, Y.}, \bibinfo{author}{{Tang}, Y.},
  \bibinfo{author}{{Xiao}, J.}, \bibinfo{author}{{Summers}, R.M.},
  \bibinfo{year}{2019}.
\newblock \bibinfo{title}{{XLSor: A Robust and Accurate Lung Segmentor on Chest
  X-Rays Using Criss-Cross Attention and Customized Radiorealistic
  Abnormalities Generation}}.
\newblock \bibinfo{journal}{arXiv e-prints} ,
  \bibinfo{pages}{arXiv:1904.09229}\href{http://arxiv.org/abs/1904.09229}{\tt
  arXiv:1904.09229}.
%Type = Article
\bibitem[{Tong et~al.(2018)Tong, Guo, Yang, Ruan and Sheng}]{tong2018}
\bibinfo{author}{Tong, N.}, \bibinfo{author}{Guo, S.}, \bibinfo{author}{Yang,
  S.}, \bibinfo{author}{Ruan, D.}, \bibinfo{author}{Sheng, K.},
  \bibinfo{year}{2018}.
\newblock \bibinfo{title}{Fully automatic multi-organ segmentation for head and
  neck cancer radiotherapy using shape representation model constrained fully
  convolutional neural network}.
\newblock \bibinfo{journal}{Med Phys} \bibinfo{volume}{45},
  \bibinfo{pages}{4558--4567}.
%Type = Article
\bibitem[{Vaassen et~al.(2020)Vaassen, Hazelaar, Vaniqui, Gooding, {van der
  Heyden}, Canters and {van Elmpt}}]{vassenAPL2020}
\bibinfo{author}{Vaassen, F.}, \bibinfo{author}{Hazelaar, C.},
  \bibinfo{author}{Vaniqui, A.}, \bibinfo{author}{Gooding, M.},
  \bibinfo{author}{{van der Heyden}, B.}, \bibinfo{author}{Canters, R.},
  \bibinfo{author}{{van Elmpt}, W.}, \bibinfo{year}{2020}.
\newblock \bibinfo{title}{Evaluation of measures for assessing time-saving of
  automatic organ-at-risk segmentation in radiotherapy}.
\newblock \bibinfo{journal}{Physics and Imaging in Radiation Oncology}
  \bibinfo{volume}{13}, \bibinfo{pages}{1 -- 6}.
\newblock \DOIprefix\doi{https://doi.org/10.1016/j.phro.2019.12.001}.
%Type = Inproceedings
\bibitem[{Vaswani et~al.(2017)Vaswani, Shazeer, Parmar, Uszkoreit, Jones,
  Gomez, Kaiser and Polosukhin}]{vaswani2017attention}
\bibinfo{author}{Vaswani, A.}, \bibinfo{author}{Shazeer, N.},
  \bibinfo{author}{Parmar, N.}, \bibinfo{author}{Uszkoreit, J.},
  \bibinfo{author}{Jones, L.}, \bibinfo{author}{Gomez, A.N.},
  \bibinfo{author}{Kaiser, {\L}.}, \bibinfo{author}{Polosukhin, I.},
  \bibinfo{year}{2017}.
\newblock \bibinfo{title}{Attention is all you need}, in:
  \bibinfo{booktitle}{Advances in Neural Information Processing Systems}, pp.
  \bibinfo{pages}{5998--6008}.
%Type = Article
\bibitem[{Vrtovec et~al.(2020)Vrtovec, Močnik, Strojan, Pernuš and
  Ibragimov}]{vrtovec2020}
\bibinfo{author}{Vrtovec, T.}, \bibinfo{author}{Močnik, D.},
  \bibinfo{author}{Strojan, P.}, \bibinfo{author}{Pernuš, F.},
  \bibinfo{author}{Ibragimov, B.}, \bibinfo{year}{2020}.
\newblock \bibinfo{title}{Auto-segmentation of organs at risk for head and neck
  radiotherapy planning: From atlas-based to deep learning methods}.
\newblock \bibinfo{journal}{Med Phys} .
%Type = Inproceedings
\bibitem[{Wang et~al.(2018a)Wang, Girshick, Gupta and He}]{wang2018non}
\bibinfo{author}{Wang, X.}, \bibinfo{author}{Girshick, R.},
  \bibinfo{author}{Gupta, A.}, \bibinfo{author}{He, K.}, \bibinfo{year}{2018}a.
\newblock \bibinfo{title}{Non-local neural networks}, in:
  \bibinfo{booktitle}{Proceedings of the IEEE Conference on Computer Vision and
  Pattern Recognition}, pp. \bibinfo{pages}{7794--7803}.
%Type = Article
\bibitem[{Wang et~al.(2019)Wang, Zhou, Shen, Park, Fishman and
  Yuille}]{WangMedIA2019}
\bibinfo{author}{Wang, Y.}, \bibinfo{author}{Zhou, Y.}, \bibinfo{author}{Shen,
  W.}, \bibinfo{author}{Park, S.}, \bibinfo{author}{Fishman, E.},
  \bibinfo{author}{Yuille, A.}, \bibinfo{year}{2019}.
\newblock \bibinfo{title}{Abdominal multi-organ segmentation with
  organ-attention networks and statistical fusion}.
\newblock \bibinfo{journal}{Medical Image Analysis} \bibinfo{volume}{55},
  \bibinfo{pages}{88--102}.
\newblock \DOIprefix\doi{10.1016/j.media.2019.04.005}.
%Type = Article
\bibitem[{Wang et~al.(2018b)Wang, Wei, Wang, Gao, Chen and
  Shen}]{wang2018hierarchical}
\bibinfo{author}{Wang, Z.}, \bibinfo{author}{Wei, L.}, \bibinfo{author}{Wang,
  L.}, \bibinfo{author}{Gao, Y.}, \bibinfo{author}{Chen, W.},
  \bibinfo{author}{Shen, D.}, \bibinfo{year}{2018}b.
\newblock \bibinfo{title}{Hierarchical vertex regression-based segmentation of
  head and neck ct images for radiotherapy planning}.
\newblock \bibinfo{journal}{IEEE Transactions on Image Processing}
  \bibinfo{volume}{27}, \bibinfo{pages}{923--937}.
%Type = Article
\bibitem[{Yuan and Wang(2018)}]{yuan2018ocnet}
\bibinfo{author}{Yuan, Y.}, \bibinfo{author}{Wang, J.}, \bibinfo{year}{2018}.
\newblock \bibinfo{title}{Ocnet: Object context network for scene parsing}.
\newblock \bibinfo{journal}{arXiv preprint arXiv:1809.00916} .
%Type = Inproceedings
\bibitem[{Zhao et~al.(2018)Zhao, Zhang, Liu, Shi, Change~Loy, Lin and
  Jia}]{zhao2018psanet}
\bibinfo{author}{Zhao, H.}, \bibinfo{author}{Zhang, Y.}, \bibinfo{author}{Liu,
  S.}, \bibinfo{author}{Shi, J.}, \bibinfo{author}{Change~Loy, C.},
  \bibinfo{author}{Lin, D.}, \bibinfo{author}{Jia, J.}, \bibinfo{year}{2018}.
\newblock \bibinfo{title}{\textsc{PSANET}: Point-wise spatial attention network
  for scene parsing}, in: \bibinfo{booktitle}{Proceedings of the European
  Conference on Computer Vision (ECCV)}, pp. \bibinfo{pages}{267--283}.
%Type = Article
\bibitem[{Zhu et~al.(2018)Zhu, Huang, Zeng, Chen, Liu, Qian, Du, Fan and
  Xie}]{zhu2018anatomynet}
\bibinfo{author}{Zhu, W.}, \bibinfo{author}{Huang, Y.}, \bibinfo{author}{Zeng,
  L.}, \bibinfo{author}{Chen, X.}, \bibinfo{author}{Liu, Y.},
  \bibinfo{author}{Qian, Z.}, \bibinfo{author}{Du, N.}, \bibinfo{author}{Fan,
  W.}, \bibinfo{author}{Xie, X.}, \bibinfo{year}{2018}.
\newblock \bibinfo{title}{Anatomynet: Deep learning for fast and fully
  automated whole-volume segmentation of head and neck anatomy}.
\newblock \bibinfo{journal}{Med Phys} \bibinfo{volume}{46},
  \bibinfo{pages}{576--589}.

\end{thebibliography}

\end{document}

\title{Appendix:\\ Nested-block self-attention for robust radiotherapy planning segmentation}

\author[mymainaddress]{Harini Veeraraghavan\corref{mycorrespondingauthor}$^{\dagger}$}
		\cortext[mycorrespondingauthor]{\\
			$^{\dagger}$ Equal contribution}
		\ead{veerarah@mskcc.org }
		\author[mymainaddress]{Jue Jiang$^{\dagger}$}
		\author[mymainaddress]{Sharif Elguindi}
		\author[mymainaddress]{Sean L. Berry} 
		\author[mysecaddress]{Ifeanyirochukwu Onochie} 
		\author[mymainaddress]{Aditya Apte} 
		\author[mymainaddress]{Laura Cervino} 
		\author[mymainaddress]{Joseph O. Deasy }
		
		\address[mymainaddress]{Department of Medical Physics,Memorial Sloan-Kettering Cancer Center, New York, USA.} 
		\address[mysecaddress]{Department of Radiation Oncology, Memorial Sloan-Kettering Cancer Center, New York, USA.}
\maketitle \vspace{-2em}
%\author{Jue Jiang\inst{1} \and Yu-Chi Hu \inst{1} \and Neelam Tyagi \inst{1} \and Perry Zhang \inst{1} \and Andreas Rimmner \inst{2} \and Gig Mageras \inst{1} \and Joseph O. Deasy \inst{1} \and Harini Veeraraghavan \inst{1}\thanks{equal contributing. Email: veerarah@mskcc.org}}

%If using runnningheads you can abbreviate the author name on even pages:
%\authorrunning{abbreviated author name}
%and you can change the author name in the table of contents
%\tocauthor{enhanced author name}

%For a single institute
%\institute{Institute Name \email{email address}}

% If authors are from different institutes 
%\institute{Medical Physics, Memorial Sloan Kettering Cancer Center \and Radiation Oncology, Memorial Sloan Kettering Cancer Center} \vspace{-3em}

%\institute{Institute:anonymous}

\section{Case by Case accuracy}

\begin{table}
\caption{\label{tab:prop_vs_manual_SDSC} SDSC of proposed method versus manual segmentation} 
    \centering
    					%\tiny
					%\scriptsize
					\footnotesize
					\small
					%\normalsize
					%\large
					%\Large
					%\LARGE
					%\huge
					%\Huge	
    {
    \begin{tabular}{|c|c|c|c|c|c|c|} \hline
         \small{Case ID} & \small{LP} & \small{RP} & \small{LS} &  \small{RS} & \small{Man} & \small{BS}\\ \hline
{0522c0017}&{0.81}&{0.88}&{0.87}&{0.76}&{0.97}&{0.97}\\
{0522c0057}&{0.92}&{0.94}&{0.95}&{0.94}&{0.96}&{0.9}\\
{0522c0161}&{0.92}&{0.84}&{0.81}&{0.8}&{0.94}&{0.98}\\
{0522c0226}&{0.97}&{0.95}&{0.93}&{0.91}&{0.96}&{0.84}\\
{0522c0248}&{0.94}&{0.89}&{0.95}&{0.92}&{0.98}&{0.95}\\
{0522c0251}&{0.95}&{0.83}&{0.95}&{0.98}&{0.95}&{0.95}\\
{0522c0331}&{0.93}&{0.84}&{0.8}&{0.89}&{0.85}&{0.9}\\
{0522c0416}&{0.91}&{0.92}&{0.95}&{0.92}&{0.87}&{0.96}\\
{0522c0419}&{0.7}&{0.72}&{0.97}&{0.94}&{0.87}&{0.92}\\
{0522c0427}&{0.94}&{0.93}&{0.85}&{0.91}&{0.89}&{0.86}\\
{0522c0457}&{0.88}&{0.95}&{0.9}&{0.91}&{0.95}&{0.96}\\
{0522c0479}&{0.94}&{0.94}&{0.78}&{0.92}&{0.92}&{0.98}\\
{0522c0629}&{0.95}&{0.98}&{0.8}&{0.9}&{0.98}&{0.98}\\
{0522c0768}&{0.94}&{0.96}&{0.9}&{0.9}&{0.97}&{0.97}\\
{0522c0770}&{0.87}&{0.91}&{0.86}&{0.87}&{0.87}&{0.95}\\
{0522c0773}&{0.92}&{0.96}&{0.97}&{0.95}&{0.88}&{0.96}\\
{0522c0845}&{0.9}&{0.86}&{0.86}&{0.93}&{0.9}&{0.95}\\
{TCGA-CV-7236}&{0.95}&{0.94}&{-}&{0.58}&{0.98}&{0.95}\\
{TCGA-CV-7243}&{0.96}&{0.93}&{0.94}&{0.59}&{0.95}&{0.95}\\
{TCGA-CV-7245}&{0.89}&{0.9}&{0.82}&{0.58}&{0.95}&{0.97}\\
{TCGA-CV-A6JO}&{0.81}&{0.8}&{-}&{-}&{0.89}&{0.98}\\
{TCGA-CV-A6JY}&{0.78}&{0.92}&{-}&{0.8}&{0.95}&{0.96}\\
{TCGA-CV-A6K0}&{0.95}&{0.95}&{0.92}&{-}&{0.98}&{0.92}\\
{TCGA-CV-A6K1}&{0.9}&{0.96}&{0.85}&{0.7}&{0.98}&{0.89}\\
\hline
    \end{tabular}
    
    }
\end{table}

\begin{table}
\caption{\label{tab:prop_vs_manual_DSC} DSC of proposed method versus manual segmentation} 
    \centering
    
    %\scriptsize
					%\tiny
					%\scriptsize
					\footnotesize
					%\small
					%\normalsize
					%\large
					%\Large
					%\LARGE
					%\huge
					%\Huge	
					
    {
    \begin{tabular}{|c|c|c|c|c|c|c|} \hline
         \small{Case ID} & \small{LP} & \small{RP} & \small{LS} &  \small{RS} & \small{Man} & \small{BS}\\ \hline
{0522c0017}&{0.74}&{0.77}&{0.73}&{0.68}&{0.92}&{0.90}\\
{0522c0057}&{0.86}&{0.86}&{0.86}&{0.83}&{0.93}&{0.86}\\
{0522c0161}&{0.85}&{0.82}&{0.77}&{0.81}&{0.91}&{0.93}\\
{0522c0226}&{0.89}&{0.88}&{0.84}&{0.81}&{0.92}&{0.84}\\
{0522c0248}&{0.81}&{0.81}&{0.88}&{0.85}&{0.95}&{0.88}\\
{0522c0251}&{0.86}&{0.80}&{0.85}&{0.88}&{0.94}&{0.90}\\
{0522c0331}&{0.84}&{0.79}&{0.76}&{0.83}&{0.87}&{0.86}\\
{0522c0416}&{0.86}&{0.85}&{0.86}&{0.86}&{0.87}&{0.89}\\
{0522c0419}&{0.72}&{0.75}&{0.91}&{0.90}&{0.90}&{0.86}\\
{0522c0427}&{0.89}&{0.86}&{0.83}&{0.86}&{0.91}&{0.87}\\
{0522c0457}&{0.83}&{0.89}&{0.87}&{0.85}&{0.91}&{0.90}\\
{0522c0479}&{0.85}&{0.86}&{0.70}&{0.84}&{0.91}&{0.90}\\
{0522c0629}&{0.87}&{0.87}&{0.77}&{0.84}&{0.92}&{0.91}\\
{0522c0768}&{0.85}&{0.87}&{0.85}&{0.83}&{0.91}&{0.90}\\
{0522c0770}&{0.82}&{0.87}&{0.80}&{0.81}&{0.88}&{0.90}\\
{0522c0773}&{0.86}&{0.88}&{0.89}&{0.88}&{0.89}&{0.90}\\
{0522c0845}&{0.85}&{0.81}&{0.82}&{0.84}&{0.90}&{0.87}\\
{TCGA-CV-7236}&{0.85}&{0.83}&{-}&{0.39}&{0.94}&{0.89}\\
{TCGA-CV-7243}&{0.88}&{0.84}&{0.87}&{0.47}&{0.91}&{0.90}\\
{TCGA-CV-7245}&{0.84}&{0.84}&{0.82}&{0.62}&{0.92}&{0.90}\\
{TCGA-CV-A6JO}&{0.81}&{0.80}&{-}&{-}&{0.88}&{0.91}\\
{TCGA-CV-A6JY}&{0.77}&{0.82}&{-}&{0.75}&{0.94}&{0.89}\\
{TCGA-CV-A6K0}&{0.89}&{0.89}&{0.84}&{-}&{0.96}&{0.89}\\
{TCGA-CV-A6K1}&{0.82}&{0.86}&{0.80}&{0.68}&{0.95}&{0.87}\\

\hline
    \end{tabular}
    
    }
\end{table}